%% file: main.tex
\documentclass[10pt,twocolumn,letterpaper]{article}

\usepackage{cvpr}              


\definecolor{cvprblue}{rgb}{0.21,0.49,0.74}
\usepackage[pagebackref,breaklinks,colorlinks,allcolors=cvprblue]{hyperref}

\usepackage{multirow}
\def\paperID{*****} 
\def\confName{CVPR}
\def\confYear{2025}

\title{Mavors: Multi-granularity Video Representation for \\ Multimodal Large Language Model}

\author{
    Yang Shi\textsuperscript{1,2}\thanks{Equal contribution.}\,\,\thanks{Work done during an internship at Kling Team.} \,
    Jiaheng Liu\textsuperscript{3}\footnotemark[1] \,
    Yushuo Guan\textsuperscript{2}\footnotemark[1] \,
    Zhenhua Wu\textsuperscript{2} \,
    Yuanxing Zhang\textsuperscript{2}\thanks{Corresponding author.} \,
    Zihao Wang\textsuperscript{2} \\
    Weihong Lin\textsuperscript{2} \,
    Jingyun Hua\textsuperscript{2} \,
    Zekun Wang\textsuperscript{2} \,
    Xinlong Chen\textsuperscript{4} \,
    Bohan Zeng\textsuperscript{1} \,
    Wentao Zhang\textsuperscript{1} \\
    Fuzheng Zhang\textsuperscript{2} \,
    Wenjing Yang \,
    Di Zhang\textsuperscript{2} \\
    \textsuperscript{1}Peking University \,
    \textsuperscript{2}Kling Team \,
    \textsuperscript{3}Nanjing University \,
    \textsuperscript{4}CASIA \\ \\
    \url{https://mavors-mllm.github.io/}
}

\begin{document}
\maketitle

\input{sec/0_abstract}
\input{sec/1_intro}
\input{sec/2_related_works}
\input{sec/3_method.tex}
\input{sec/4_training_paradigm}
\input{sec/5_experiments}
\input{sec/6_conclusion}
{
    \small
    \bibliographystyle{ieeenat_fullname}
    \bibliography{main}
}

\input{sec/X_suppl}

\end{document}

%% file: sec/0_abstract.tex
\begin{abstract}
    Long-context video understanding in multimodal large language models (MLLMs) faces a critical challenge: balancing computational efficiency with the retention of fine-grained spatio-temporal patterns. Existing approaches (e.g., sparse sampling, dense sampling with low resolution, and token compression) suffer from significant information loss in temporal dynamics, spatial details, or subtle interactions, particularly in videos with complex motion or varying resolutions. To address this, we propose \textbf{Mavors}, a novel framework that introduces \textbf{M}ulti-gr\textbf{a}nularity \textbf{v}ide\textbf{o} \textbf{r}epre\textbf{s}entation for holistic long-video modeling. 
    Specifically,
    Mavors directly encodes raw video content into latent representations through two core components: 1) an \textbf{Intra-chunk Vision Encoder (IVE)} that preserves high-resolution spatial features via 3D convolutions and Vision Transformers, and 2) an \textbf{Inter-chunk Feature Aggregator (IFA)} that establishes temporal coherence across chunks using transformer-based dependency modeling with chunk-level rotary position encodings. Moreover, the framework unifies image and video understanding by treating images as single-frame videos via sub-image decomposition.
    Experiments across diverse benchmarks demonstrate Mavors' superiority in maintaining both spatial fidelity and temporal continuity, significantly outperforming existing methods in tasks requiring fine-grained spatio-temporal reasoning.
\end{abstract}

%% file: sec/1_intro.tex
\section{Introduction}
\label{sec:intro}

\begin{figure*}[!htp]
  \centering
  \includegraphics[width=0.9\linewidth]{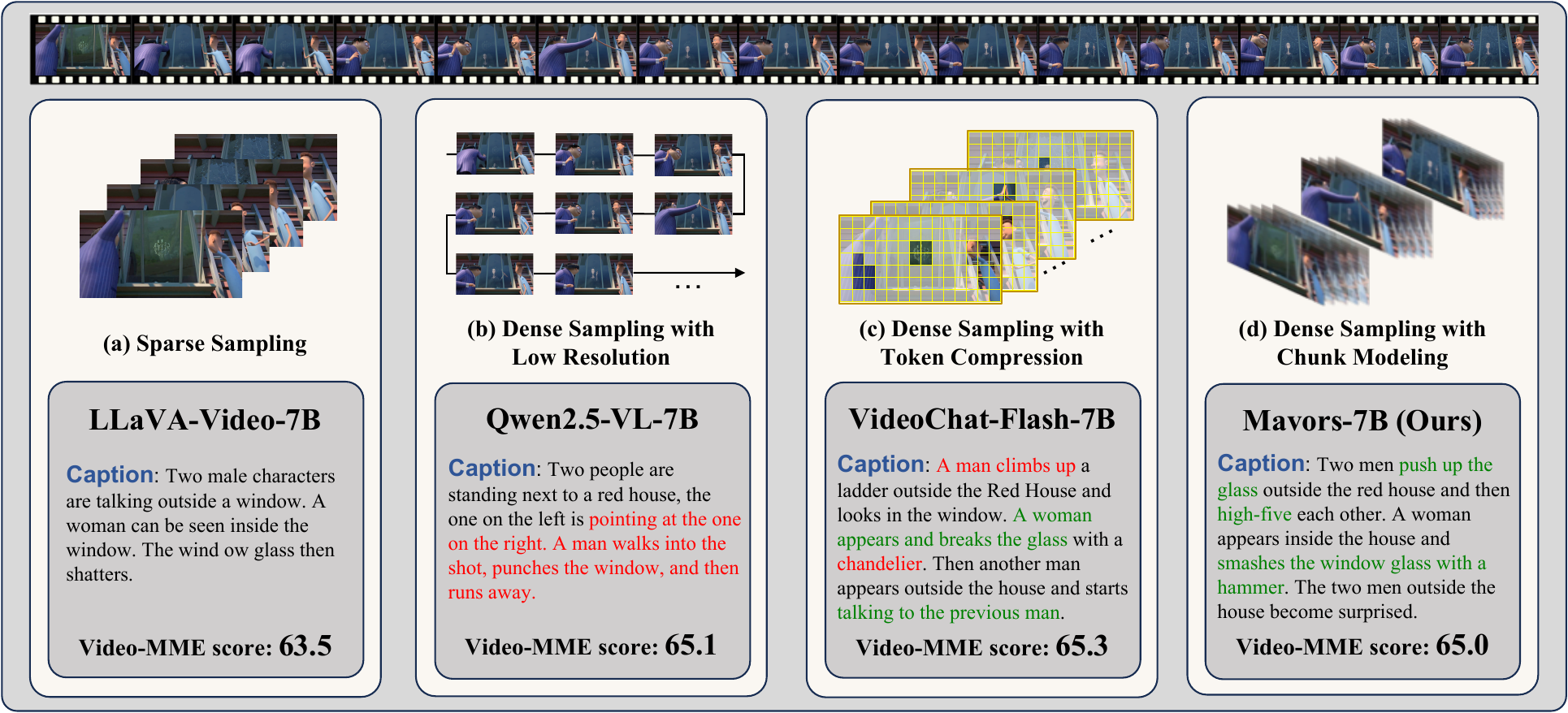}
  \caption{(a) Sparse sampling, which remains the high resolution but loses many details in the unsampled frames; (b) Dense sampling with low resolution, which understands the videos from a large number of frames but would confuse on the low-resolution content; (c) Dense sampling with token compression, which keeps the key tokens on the main characters but suffers from hallucinations owing to the missing of visual tokens; (d) Our Mavors, balancing the demands of resolution and number of frames. Though all these approaches could perform similarly on Video-MME, Mavors significantly improves the caption capability on complex scenes. Note that the words in red and green denote incorrect and correct details, respectively.}\label{fig:intro}
\end{figure*}
Long-context video modeling stands as one of the most crucial capabilities within MLLMs~\cite{fuyu-8b,zhang2025videollama3, videochat-flash, Maaz2023VideoChatGPT}. This capability empowers MLLMs to proficiently manage hours-long movies, documentaries, and online video streams, all of which demand sophisticated long video processing. Recent advances in MLLMs perform well in short video understanding. However, it remains challenging to build MLLMs for processing extremely long videos (lasting for hours or even longer). The difficulty lies in how to enable MLLMs to efficiently understand the extremely long video context brought by long videos.

As shown in Figure~\ref{fig:intro}, 
we have compared three mainstream types of video MLLMs with our method,
and provided the video caption results of different methods for better illustration.
Specifically, 
in Figure~\ref{fig:intro}(a),
these methods (e.g., LLaVA-Video~\cite{zhang2024video}, InternVL 2.5~\cite{chen2024expanding}) usually employ the sparse sampling strategy to decrease the number of frames and reduce the computation costs. However, these methods have a significant limitation, where many temporal contexts are lost as many frames are not sampled. Thus, the performance results of video-related tasks, which require detailed temporal contexts from many frames, are degraded a lot for these methods.
When compared to methods in Figure~\ref{fig:intro}(a),
some methods (e.g., Oryx~\cite{liu2024oryx}, Qwen2VL~\cite{Qwen2VL}) have introduced the strategy of dense sampling with low-resolution input in Figure~\ref{fig:intro}(b). However, for these methods, many spatial contexts are lost as only the low-resolution frames are given, which also significantly degrade the results of video-related tasks requiring detailed spatial contexts, e.g., video captioning.
Recently, 
in Figure~\ref{fig:intro}(c),
several works (e.g., VideoLLaMA 3~\cite{zhang2025videollama3}, VideoChat-Flash~\cite{videochat-flash}) have proposed token compression strategies (e.g., token merge or token dropping), which reduces tokens based on vector or pixel similarity and effectively preserves spatial-temporal features of large visual elements.
However, token compression inevitably leads to the loss of information regarding small spatial objects, subtle temporal motions, and interactions among multiple objects, thereby posing challenges for understanding complex scenes.

Therefore, the fundamental problem of video understanding is that
\textbf{existing methods often rely on sparse sampling or token compression strategies and struggle to balance computational efficiency with the retention of fine-grained spatio-temporal patterns, particularly in videos with variable motion, aspect ratios, or resolutions.}

To address this problem,
as shown in Figure~\ref{fig:intro}(d),
we introduce the \textbf{Mavors} method to extract the \textbf{M}ulti-gr\textbf{a}nularity \textbf{v}ide\textbf{o} \textbf{r}epre\textbf{s}entation for MLLMs.
which is designed to process raw video content holistically while preserving both spatial fidelity and temporal coherence.  
Specifically,
 Mavors eliminates the information loss inherent in conventional frame sampling or token compression methods by directly encoding consecutive video chunks into latent representations. This approach leverages a two-tier architecture: an \textbf{Intra-chunk Vision Encoder (IVE)} extracts high-resolution spatial features from localized video segments using 3D convolutions and Vision Transformer (ViT) layers, while an \textbf{Inter-chunk Feature Aggregator (IFA)} employs temporal transformer and chunk-level rotary position embeddings (C-RoPE) to model temporal dependencies across chunks. 
Besides,
Mavors further unifies image and video understanding by treating images as single-frame videos by employing a sub-image divide-and-conquer approach for image processing.
Moreover, following the common training strategy,
we also adopt a multi-stage training paradigm,
which includes the modality alignment, temporal understanding enhancement, instruction tuning and DPO training stages.

The contributions of Mavors are shown as follows:
\begin{itemize}
\item We propose the \textbf{Mavors} by utilizing the \textbf{M}ulti-gr\textbf{a}nularity \textbf{v}ide\textbf{o} \textbf{r}epre\textbf{s}entation for multimodal large language model, which aims to better preserve the spatio-temporal contexts based on dense sampling with chunk modeling. 
\item Mavors includes two modules: \textbf{Intra-chunk Vision Encoder (IVE)} and \textbf{Inter-chunk Feature Aggregator (IFA)}. IFA encodes consecutive video chunks into latent representation based on 3D convolutions and ViT, and IFA builds the temporal coherence based on the temporal transformer and chunk-level rotary-encoding strategies.   
\item Comprehensive experimental results and detailed analysis show the effectiveness and efficiency of Mavors.
\end{itemize}

%% file: sec/2_related_works.tex
\section{Related Works}
\label{sec:formatting}

\subsection{MLLM Architecture}
Current MLLMs employ two architectural strategies for visual processing.
The first paradigm is based on cross-attention approach, which maintains frozen model parameters while establishing dynamic visual-language interactions through attention mechanisms~\cite{Alayrac2022FlamingoAV} . Alternatively, the second paradigm processes visual content through pretrained encoders (CLIP~\cite{radford2021clip}, SigLIP~\cite{siglip}) before concatenating image tokens with text embeddings for unified language model processing~\cite{lin2023vila,liu2023llava,liu2023llava15,li2024llava,liu2024llavanext}.
The second paradigm can be readily extensible to video analysis through sequential frame processing~\cite{2023videochat,zhang2025videollama3},
and many architectural innovations for temporal modeling have been proposed~\cite{Jiang2025TokenEfficientLV, wang2025internvideo2,kangaroogroup}.

\subsection{MLLM for Video Understanding}
Existing MLLMs have revealed divergent capabilities in temporal comprehension across different video durations. While existing systems demonstrate proficiency in minute-scale video analysis~\cite{2023videochat, videochat-flash, lin2023video}, emerging efforts targeting hour-level sequences~\cite{gemini,wang2024longllava} face fundamental challenges. 
To address the challenges of long video modeling,
current approaches primarily pursue two optimization directions: (1) context window expansion for large language models~\cite{gemini,wang2024longllava,Zhang2024LongCT, longvila} and (2) efficient token compression via spatial-temporal feature distillation~\cite{videoccam, li2024llamavid, shu2024video, song2023moviechat, TanKoala2024, Weng2024LongVLMEL}. 
For the first 
strategy, though theoretically enabling long-sequence processing, suffers from impractical computational overhead, which bring significant challenges for practical applications.
In contrast, 
recent token compression methods like LLaMA-VID~\cite{li2024llamavid} achieve compression rates at the cost of discarding subtle details,
which results in performance degradation on standard video understanding benchmarks. 
When compared to the existing works, our Mavors can directly process the raw videos to maintain spatial and temporal details well with acceptable computation costs. 

%% file: sec/3_method.tex
\section{Method}
\label{sec:method}

\subsection{Preliminaries}

\begin{figure}[t]
  \centering
  \begin{subfigure}[t]{0.44\linewidth}
    \centering
    \includegraphics[width=\linewidth]{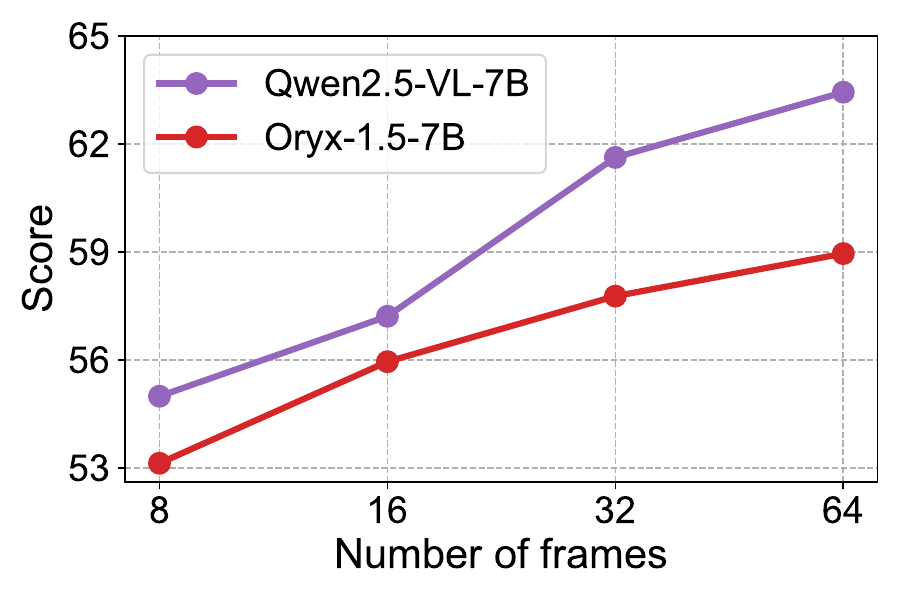}
    \vspace{-1.5em}
    \caption{Video-MME}
    \label{fig:fps-Video-MME1}
  \end{subfigure}
  \hfill
  \begin{subfigure}[t]{0.44\linewidth}
    \centering
    \includegraphics[width=\linewidth]{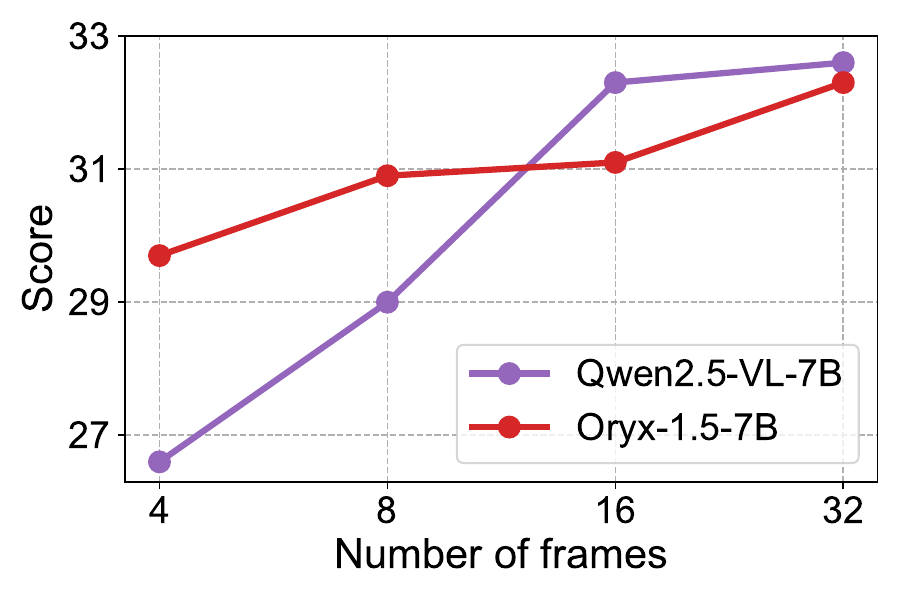}
    \vspace{-1.5em}
    \caption{Dream1K}
    \label{fig:fps-dream1k1}
  \end{subfigure}
  \vspace{-1em}
  \caption{The impact of the number of frames (720P).}
  \label{fig:fps_score}
\end{figure}

\begin{figure}[t]
  \centering
  \begin{subfigure}[t]{0.44\linewidth}
    \centering
    \includegraphics[width=\linewidth]{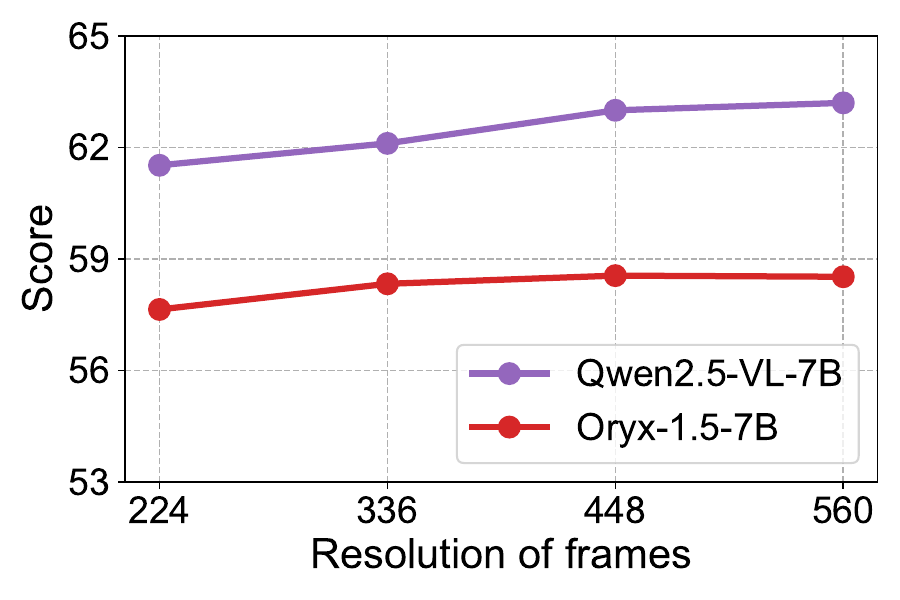}
    \vspace{-1.5em}
    \caption{Video-MME}
    \label{fig:fps-Video-MME2}
  \end{subfigure}
  \hfill
  \begin{subfigure}[t]{0.44\linewidth}
    \centering
    \includegraphics[width=\linewidth]{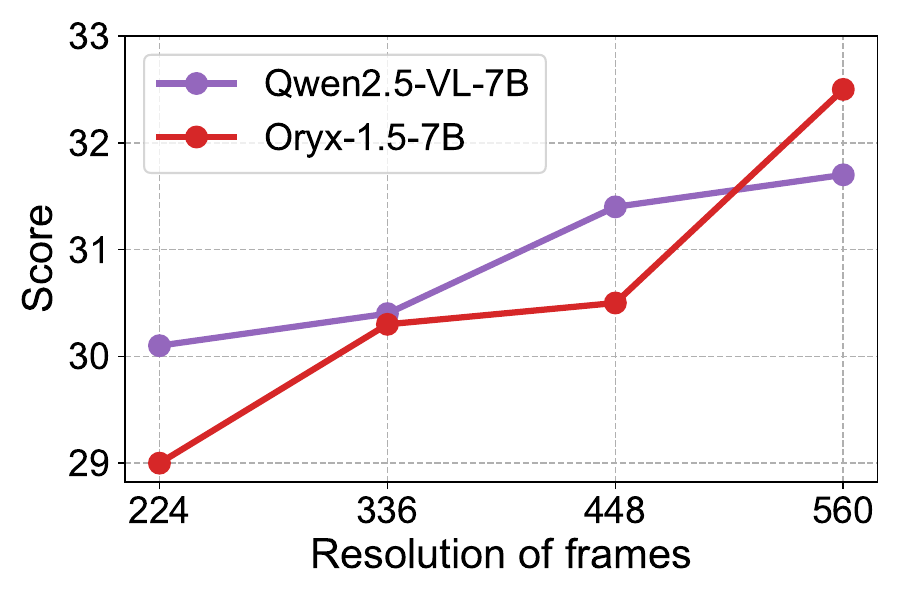}
    \vspace{-1.5em}
    \caption{Dream1K}
    \label{fig:fps-dream1k2}
  \end{subfigure}
  \vspace{-1em}
  \caption{The impact of the resolution of frames (64 frames).}
  \label{fig:res_score}
\end{figure}

\noindent\textbf{Necessity of Dense Sampling with High Resolution}.
As shown in Figure~\ref{fig:fps_score} and Figure~\ref{fig:res_score},
we have compared the results of two popular video MLLMs (i.e., Qwen2.5-VL-7B~\cite{qwen2.5vl} and Oryx-1.5-7B~\cite{liu2024oryx}) on two representative benchmarks (i.e.,  Video-MME~\cite{videomme} and DREAM-1K~\cite{wang2024tarsier}).
Specifically,
the Video-MME focuses on multiple-choice question answering based on video content and requires a better understanding of the temporal relations between different frames.
DREAM-1K involves open-ended video captioning, where models must generate detailed descriptions of the main events in the video.
Thus, both the spatial and temporal fine-grained details are important. In Figure~\ref{fig:fps_score},
we observe that performance increases a lot when increasing the number of frames, which shows the necessity of dense sampling with more frames.
In Figure~\ref{fig:res_score},
performance results on Video-MME are relatively stable for both MLLMs. For this phenomenon, we assume that understanding fine spatial details is not vital for Video-MME.
In contrast, the results on DREAM-1K increase a lot, which demonstrates the necessity of high resolution.

In summary, as real-world video understanding tasks usually rely on understanding the fine-grained spatiotemporal contexts well, it is important to design video MLLMs by sampling dense and high-resolution frames and maintaining efficiency.

\begin{figure*}[ht]
  \centering
  \includegraphics[width=0.8\linewidth]{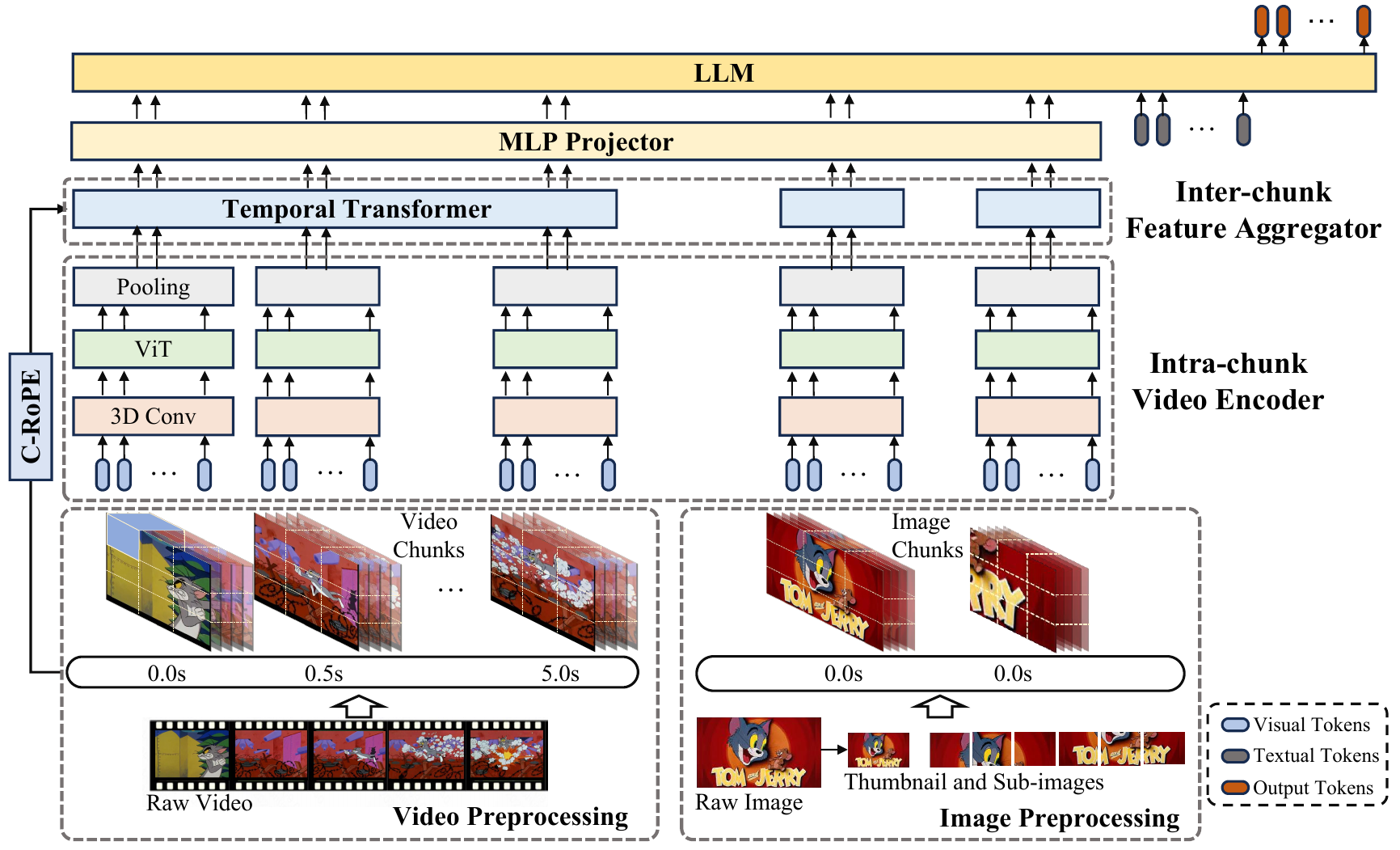}
  \caption{The architecture of Mavors.}\label{fig:model_arch}
\end{figure*}

\subsection{Overview of Mavors}

In Figure~\ref{fig:model_arch},
the key objective of Mavors is to enhance the video understanding capability by introducing an efficient video encoding strategy based on dense sampling with high resolution strategy. 

Specifically, Mavors employs a video encoder that directly processes pixel information from video chunks, converting them into latent representations. 
Figure~\ref{fig:model_arch} illustrates the overview of Mavors when dealing with video content and images.
We consider an input video $S_\text{V}\in \mathbb{R}^{W_\text{V}\times H_\text{V}\times 3\times T_\text{V}}$ or an image $S_\text{I}\in \mathbb{R}^{W_\text{I}\times H_\text{I}\times 3}$, where $W_\text{V}, H_\text{V}$ and $W_\text{I}, H_\text{I}$ denote the respective widths and heights, and $T_\text{V}$ denotes the total number of video frames.
Mavors follows the auto-regressive architecture to generate a textual response based on a given textual instruction.
Specifically, in Mavors, we first perform the \textbf{preprocessing}  on the raw videos or images to obtain the model input.
Then, we employ an \textbf{intra-chunk vision encoder} and an  \textbf{inter-chunk feature aggregator} to fully comprehend videos, so that the spatial and temporal details would be remained. 
Following the mainstream architecture of MLLMs, the temporally integrated features are passed through an MLP projector for modality alignment before being input to the LLM.


\subsection{Intra-chunk Vision Encoder}

Mavors partitions the video frames into $c_\text{V}=\lceil\frac{T_\text{V}}{F}\rceil$ video chunks, where each chunk contains $F$ consecutive frames describing the dynamic scenes and temporal events, i.e.,
$C_{1,\ldots,c_\text{V}} = \text{Partition}(S_\text{V})$.
Intra-chunk vision encoder is designed to represent the vision features of the video content.
It begins with 3D convolutions applied to individual video chunks, and we would obtain the visual feature $\mathcal{F}_{i}$ for the $i$-th chunk as follows:
\begin{equation}\label{eqn:conv}
\mathcal{F}_{i} = \text{Conv}(C_i)/F\in\mathbb{R}^{n_\text{V}\times d_{\text{V}}}, i=1,\ldots, c_\text{V},
\end{equation}
where $n_\text{V}$ indicates the number of visual features per video chunk, and $d_{\text{V}}$ denotes the dimension of the visual features.
We then adopt a standard ViT with parameter $\theta_{\text{ViT}}$ to capture high-level spatial-temporal features, denoted as $\hat{\mathcal{H}}_{i}$, within the $i$-th chunk.
To manage the computational load and complexity for the downstream LLM module arising from a large number of tokens, we apply a 2x2 pooling layer on $\hat{\mathcal{H}}_{i}$ to obtain $\mathcal{H}_{i}\in\mathbb{R}^{n_\text{V}/4\times d_{\text{V}}}$.

We initialize $\theta_{\text{ViT}}$ by SigLIP weights.
Specifically, the 2D convolutional kernels from SigLIP are replicated $F$ times along the temporal dimension to form the 3D kernels.
As the resulting visual features are divided by $F$ in Eqn. (\ref{eqn:conv}), the spatial absolute position embedding is added to the feature vectors towards the corresponding pixel patches.
This ensures that the model's initial behavior precisely matches its capability for single image-text understanding. 


\subsection{Inter-chunk Feature Aggregator}

The intra-chunk vision encoder mainly captures the high-level visual features within video chunks.
Mavors leverages the the inter-chunk feature aggregator, to integrate temporal information across the multiple video chunks of the complete video. 
First, we concatenate the high-level visual features to form the original feature sequence as follows:
\begin{equation}
\chi^{(0)} = \text{Concat}(\mathcal{H}_{1, \ldots, c_\text{V}}).
\end{equation}
Inter-chunk feature aggregator consists of $L_\text{inter}$ Transformer layers with Causal Attention. 
To identify the sequential order of the visual features, we propose \emph{chunk-level Rotary Encoding} (C-RoPE) to the Transformer layers, so that the temporal information can be correctly retained.
Specifically, the causal scaled dot product (SDP) attention in the $j$-th Transformer layer would be calculated by
\begin{equation}
\mathcal{Q}_{\text{Inter}}^{(j)}, \mathcal{K}_{\text{Inter}}^{(j)},\mathcal{V}_{\text{Inter}}^{(j)} = \text{Linear}(\chi^{(j-1)}),
\end{equation}
\begin{equation}
\begin{aligned}
\text{SDP}(q_{\iota}^{(j)},k_{\iota'}^{(j)}) &= \text{C-RoPE}(q_{\iota}^{(j)}, k_{\iota'}^{(j)}; \lceil\frac{4\iota}{n_{\text{V}}}\rceil, \lceil\frac{4\iota'}{n_{\text{V}}}\rceil) \\
&=q_{\iota}^{(j)}R_{\lfloor\frac{4\iota}{n_{\text{V}}}\rfloor-\lfloor\frac{4\iota'}{n_{\text{V}}}\rfloor}k_{\iota'}^{(j)\intercal}, \\
& \forall q_{\iota}^{(j)}\in \mathcal{Q}_{\text{Inter}}^{(j)}, k_{\iota'}^{(j)}\in\mathcal{K}_{\text{Inter}}^{(j)}
\end{aligned}
\end{equation}
Here, $R$ represents the rotation matrix.
In practice, we would transcode the video into fixed FPS, so that the index of the video chunk can be identified from the actual timestamp of the first frame of the chunk.
In the remaining process of the Transformer layer, we follow
\begin{equation}
\mu^{{j}} = \text{softmax}(\text{SDP}(\mathcal{Q}_{\text{Inter}}^{(j)}, \mathcal{K}_{\text{Inter}}^{(j)})),
\end{equation}
\begin{equation}
\chi^{(j)} = \mu^{{j}}\mathcal{V}_{\text{Inter}}^{(j)}.
\end{equation}
We then feed $\chi^{(L_\text{Inter})}$ to the MLP projector to obtain the visual tokens,
where the feature dimension of these visual tokens is the same as the feature dimension of  textual tokens in LLM.


\begin{figure}[ht]
  \centering
  \includegraphics[width=0.9\linewidth]{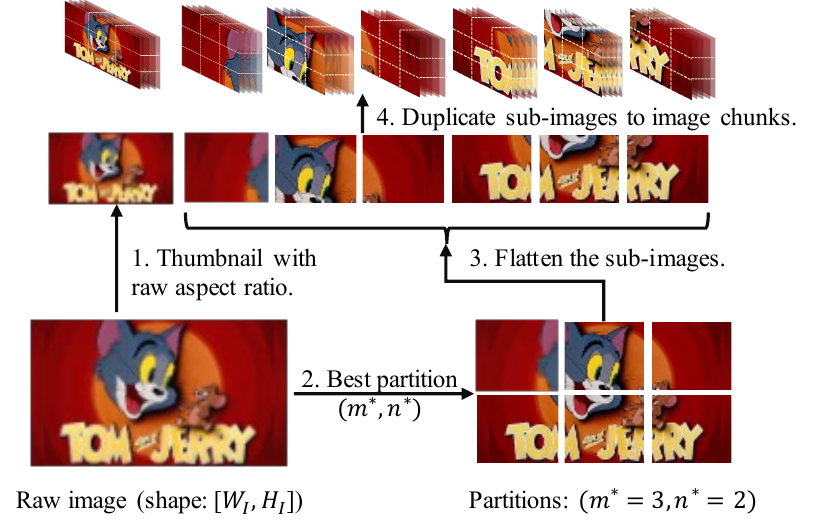}
  \caption{The dynamic resolution strategy in Mavors.}\label{fig:tiling}
\end{figure}

\begin{figure*}[t]
  \centering
  \includegraphics[width=0.8\linewidth]{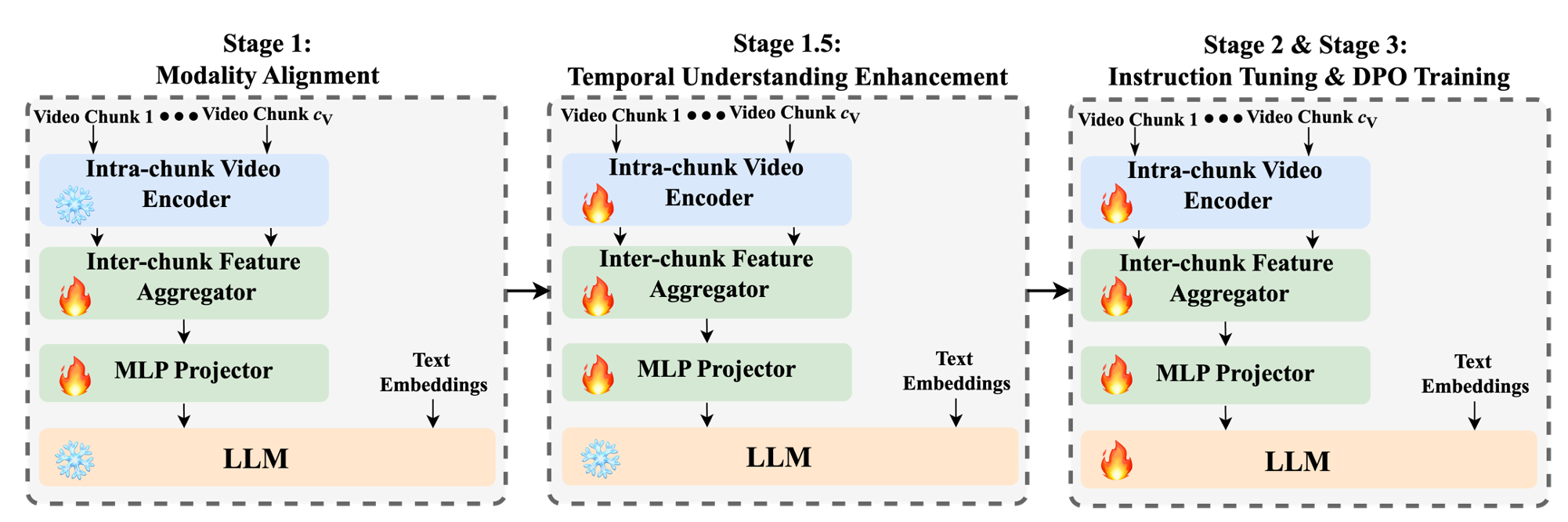}
  \caption{Training paradigm of different stages.}\label{fig:stages}
\end{figure*}
\subsection{Preprocessing}

\noindent\textbf{Video Preprocessing.}
The video processing strategy of Mavors varies based on the video length.
Specifically,
videos with short lengths are directly processed into chunks.
To accommodate long videos, we employ an initial step of accelerated playback achieved through frame dropping, thereby reducing the total frame count to be compatible with Mavors processing limits.
Specifically, the position IDs utilized by C-RoPE correspond to timestamps derived from the original, non-accelerated video timeline.
This mechanism informs the model that the processed frames are not temporally contiguous.
While alternative strategies for very long video comprehension exist, e.g., in-video Retrieval-Augmented Generation (RAG)~\cite{luo2024video}, they represent an orthogonal direction to Mavors.

Meanwhile,
Mavors could process videos with arbitrary resolutions and aspect ratios. Specifically, Mavors employs a dynamic resolution strategy to maintain the original aspect ratio of the video frames, avoiding distortion artifacts that can arise from fixed-shape resizing. The resized video frames roughly keep the original aspect ratio and match the number of pixels in the ViT's pretraining images. For example, given the frames with the $(W_\text{V}, H_\text{V})$ resolution and the ViT's pretrained image resolution $(R_v, R_v)$, Mavors will rescale the frames into the resolution of $(R_v * \sqrt{W_\text{V}/H_\text{V}}, R_v * \sqrt{H_\text{V}/W_\text{V}})$. We also resize the positional embedding of patches, following SigLIP~\cite{siglip}. Specifically, the positional embedding of the video chunk in the $(x, y)$ position, denoted as $E(x,y)$, will be formulated as:
\begin{equation}
E(x,y)=E_v(x * P_v / P_W, y * (P_v / P_H)),
\end{equation}
where $(P_W, P_H)$ is the number of patches in the video chunk. $P_v$ and $E_v(x,y)$ are the number of patches and the positional embedding in the ViT's pretraining images, respectively.

\noindent\textbf{Image Preprocessing.}
As shown in Figure~\ref{fig:tiling},
Mavors first partitions the raw image into several sub-images, and then leverages the thumbnail of the original image and all sub-images into the vision encoder. Besides, Mavors incorporates a special design in the feature aggregator to accommodate the joint training of videos and images. 
The details are as follows.

First,
as image understanding tasks often require spatial details, we follow the image partition method in~\cite{yao2024minicpmvgpt4vlevelmllm} and support dynamic resolution for processing high-resolution images,
where the raw image will be partitioned into multiple sub-images and the size of these sub-images is supposed to match the number of pixels in the ViT's pretraining.
Specifically, we first determine the ideal number of sub-images $N_s = \left\lfloor (W_\text{I} \times H_\text{I})/R_v^2 \right\rfloor$, where ($W_\text{I}, H_\text{I}$) is the resolution of the original raw image and ($R_v, R_v$) is the resolution of the ViT's pretraining images. 
Next, we identify potential partition configurations by finding pairs of integers $(m, n)$, representing the number of columns and rows, respectively, such that their product equals the target number of slices $N_s$. These pairs form the set $\mathcal{C}_{N_s} = \{(m, n) | m \times n = N_s, m, n \in \mathbb{Z}\}$. 
Then, we select the best configuration $(m^*, n^*)$ from $\tilde{C} = \mathcal{C}_{N_s-1} \cup \mathcal{C}_{N_s} \cup \mathcal{C}_{N_s+1}$ based on the following criteria:
\begin{equation}
  \quad (m^*, n^*) = \arg \min_{(m,n) \in \tilde{C}} \left| \log \frac{W_\text{I}}{H_\text{I}} - \log \frac{m}{n} \right|.
\end{equation}
We will leverage the thumbnail of the original raw image $I_0$ and all sub-images $I_1, ..., I_{m^*\times n^*}$ as the input of the vision encoder. Before feeding into the vision encoder, we will rescale the original image and the sub-images, which have more pixels than the ViT's pretraining images. We use the same dynamic resolution strategy as video processing.


Second,
when compared to video processing,
the feature aggregator operates on the features extracted from each sub-image independently, thus avoiding redundant temporal relationships.
Furthermore, given that the model must process both images and videos, the representation of an image (treated as a single frame) is replicated across all temporal positions within the input sequence.
Placing the image representation at only a single temporal position would
cause the model parameters to become biased towards that static position, ultimately hindering the model's capacity to perceive temporal information effectively in video sequences.


%% file: sec/4_training_paradigm.tex
\section{Training Paradigm}
\label{sec:training_paradigm}

In Figure~\ref{fig:stages},
multi-stage training is adopted, serving to improve the collaboration of the video encoder and LLM and the performance of multimodal tasks.
Given SigLIP's robust image understanding performance, we forgo an independent CLIP training phase to avoid redundancy. Instead, we adopt a tailored initialization strategy to ensure compatibility with both video and image inputs, where the 2D convolutional kernels from SigLIP are replicated $F$ times along the temporal dimension to form the 3D kernels.
Then, we leverage multiple training stages to progressively build a vision encoder that maintains image understanding while effectively encoding spatio-temporal information of videos.
The data used for training Mavors is detailed in Appendix A.

\noindent\textbf{Stage 1: Modality Alignment.}
As SigLIP's training involved alignment with the T5 model~\cite{2020t5}, the first stage aims to align the semantic space of the vision encoder with the LLM's semantic space. 
In this stage, we train the inter-chunk feature aggregator and the MLP projector, while keeping the LLM and the intra-chunk vision encoder frozen. 
Although the model exhibits only coarse video comprehension at this stage, the principal aim is to achieve modality alignment and instill basic temporal understanding.
Therefore, we prioritize diverse, general-concept image-text pairs and short video-text pairs with low complexity (e.g., LAION~\cite{laion400m} and PANDA-70M\cite{chen2024panda}), thereby avoiding excessively difficult data that could impede the development of foundational abilities.

\noindent\textbf{Stage 1.5: Temporal Understanding Enhancement.}
Subsequent to Stage 1, we implement Stage 1.5, which focuses on enhancing the video encoder's capacity for genuine video comprehension.
Based on the modality alignment from Stage 1, parameter updates are performed on all components excluding the LLM.
For data selection in this stage, we augment the initial dataset with standard computer vision (CV) tasks applied to images and short video chunks, such as captioning, classification, OCR, interleaved image-text, and perception QA.

\noindent\textbf{Stage 2: Multitask Instruction Tuning.}
In Stage 2, the primary objective is to adapt the model for a range of multi-modal tasks, leveraging data formats including text-only, single-image, multi-images, and complex video.
Beyond standard CV tasks, we incorporate grounding tasks and temporal grounding tasks to enhance the model's perception of spatio-temporal details. 
Similar to the practice in Qwen2.5VL~\cite{qwen2.5vl}, we find that representing bounding boxes using plain text coordinates yields performance comparable to using special tokens; consequently, we adopt the plain text representation.
This stage also activates the sub-image partitioning paradigm to enhance the model's image understanding capabilities.
All model parameters are unfrozen and trained on a large dataset, allowing for extensive self-adjustment.
Upon completion, the model possesses significant world knowledge, semantic understanding, and logical reasoning abilities, though its application is initially limited by the specific tasks and query formats encountered.
Therefore, towards the end of this stage, we introduce more diverse data types, covering a broader spectrum of real-world task scenarios and textual query formulations.

\noindent\textbf{Stage 3: DPO Training.}
Our empirical evaluations reveal that while the previously described training procedure yields strong leaderboard performance, the resulting model exhibits distinct patterns.
Specifically, for QA tasks, the model tends to generate overly concise responses, likely due to extensive training on multiple-choice or short-answer datasets.
Conversely, for descriptive tasks, the model fails to terminate generation appropriately.
To mitigate these issues, we incorporate a Direct Preference Optimization (DPO) \cite{Rafailov2023DirectPO} stage following Stage 2.
The preference dataset  mainly covers three domains: open-ended QA, image captioning, and video captioning.
More details can be found in Appendix A.

\noindent\textbf{Loss Function.}
We employ the next-token-prediction (NTP) training methodology in all training stages except the DPO stage.
During DPO training, we employ the standard DPO loss.


%% file: sec/5_experiments.tex
\section{Experiments}
\label{sec:experiments}

\begin{table*}[t]
    \centering
    \resizebox{\linewidth}{!}{
    \begin{tabular}{lccccccccccc}
    \toprule
        \textbf{Model} & \textbf{Size} & \textbf{MMWorld} & \textbf{PerceptionTest} & \textbf{Video-MME} & \textbf{MLVU} & \textbf{MVBench} & \textbf{EventHallusion} & \textbf{TempCompass} & \textbf{VinoGround} & \textbf{DREAM-1K} \\
        \midrule
        \textcolor{lightgray}{GPT-4o-20240806} & \textcolor{lightgray}{-} & \textcolor{lightgray}{62.5} & \textcolor{lightgray}{-} & \textcolor{lightgray}{71.9} & \textcolor{lightgray}{64.6} & \textcolor{lightgray}{64.6} & \textcolor{lightgray}{92.0} & \textcolor{lightgray}{73.8} & \textcolor{lightgray}{38.9} & \textcolor{lightgray}{39.2} \\
          \textcolor{lightgray}{Gemini-1.5-Pro} & \textcolor{lightgray}{-} & \textcolor{lightgray}{-} & \textcolor{lightgray}{-} & \textcolor{lightgray}{75.0} & \textcolor{lightgray}{-} & \textcolor{lightgray}{60.5} & \textcolor{lightgray}{80.3} & \textcolor{lightgray}{67.1} & \textcolor{lightgray}{22.9} & \textcolor{lightgray}{36.2} \\
                \midrule
        LLaVA-OneVision & 7B & 59.2 & 56.9 & 58.9 & 64.8 & 56.7 & 64.3 & 61.4 & 26.2 & 31.9 \\
        InternVL 2.5 & 8B & 62.2 & 65.0 & 64.3 & 67.0 & 72.0 & 64.1 & 71.4 & 24.0 & 29.7 \\
        NVILA & 8B & 55.2 & 55.5 & 64.2 & 70.1 & 68.1 & 69.9 & 66.5 & 20.2 & 26.9 \\
        LLaVA-Video & 7B & 60.1 & 67.5 & 63.6 & 67.2 & 58.6 & 70.7 & 65.7 & 26.9 & 33.3 \\
        Oryx-1.5 & 7B & 58.8 & 70.3 & 59.0 & 63.8 & 67.5 & 61.3 & 60.2 & 22.3 & 32.5 \\
        Qwen2.5-VL & 7B & 61.3 & 66.2 & 65.1 & 70.2 & 69.6 & 66.5 & 71.4 & 34.6 & 32.6\\
        VideoLLaMA3 & 7B & 56.4 & 72.8 & \textbf{66.2} & 73.0 & 69.7 & 63.4 & 68.1 & 31.3 & 30.5 \\
        VideoChat-Flash & 7B & 57.9 & \textbf{74.7} & 65.3 & \textbf{74.7} & \textbf{74.0} & 66.4 & 70.0 & 33.3 & 29.5\\
        Slow-fast MLLM & 7B & 58.2 & 69.7 & 60.2 & 60.4 & 68.9 & 67.4 & 69.9 & 27.1 & 33.2 \\
        \midrule
        \textcolor{lightgray}{Qwen2.5-VL} & \textcolor{lightgray}{72B} & \textcolor{lightgray}{73.1} & \textcolor{lightgray}{73.2} & \textcolor{lightgray}{73.3} & \textcolor{lightgray}{76.6} & \textcolor{lightgray}{70.4} & \textcolor{lightgray}{76.3} & \textcolor{lightgray}{79.1} & \textcolor{lightgray}{58.6} & \textcolor{lightgray}{35.1} \\
        \textcolor{lightgray}{InternVL 2.5} & \textcolor{lightgray}{78B} & \textcolor{lightgray}{77.2} & \textcolor{lightgray}{73.5} & \textcolor{lightgray}{72.1} & \textcolor{lightgray}{76.6} & \textcolor{lightgray}{76.4} & \textcolor{lightgray}{67.7} & \textcolor{lightgray}{75.5} & \textcolor{lightgray}{38.7} & \textcolor{lightgray}{30.3}  \\
        \midrule
        Mavors (Ours) & 7B & \textbf{68.1} & 70.3 & 65.0 & 69.8 & 68.0 & \textbf{73.5} & \textbf{77.4} & \textbf{36.9} & \textbf{39.4} \\
        \bottomrule
    \end{tabular}
    }
    \caption{Performance on video benchmarks. Most of the scores are from their original studies. The others are reproduced following the official benchmark recommendation.}
    \label{tab:video}
    \vspace{-4mm}
\end{table*}

\subsection{Experimental Setup}
\noindent\textbf{Implementation Details}.
The Mavors model utilizes Qwen2.5-7B as its language model module, with the intra-chunk vision encoder initialized using SigLIP weights.
To balance effectiveness and efficiency, the frame count per video chunk, $F$, is set to 16.
The inter-chunk feature aggregator consists of $L_\text{Inter}$=3 layers.
The training is conducted on 416 GPUs.
Given the model's moderate size, we employed DeepSpeed with ZeRO stage 2 optimization.
As mentioned in Section~\ref{sec:training_paradigm}, the pre-training proceeded in three stages: 
Stage 1 used approximately 127 million samples with a global batch size of 6,656, taking 71 hours;
Stage 1.5 used 52 million samples with a global batch size of 3,328, taking 177 hours;
and Stage 2 used 19 million samples with a global batch size of 1,664, requiring 28 hours.
The learning rates for the LLM and projector are set to 1e-5 in both Stage 1 and Stage 1.5, with a constant learning rate schedule applied during these phases. In Stage 2 and DPO, the learning rate was initialized at the same value (1e-5) as the preceding stages but followed a cosine decay schedule, gradually reducing to 1/10th of its initial value. Meanwhile, the learning rates for the inter-chunk feature aggregator and intra-chunk vision encoder remained fixed at 1/10th of the LLM's learning rate across all training stages.

For inference, Mavors is adapted using the vLLM framework~\cite{kwon2023efficient}.
Since Mavors requires comprehensive video encoding and frame preprocessing occurs on the CPU, the CPU processor can thus become a bottleneck. 
Recognizing that the intra-chunk vision encoder's computation is a one-time GPU operation per video, with results stored in the LLM's KV cache, we overlaps the pipeline.
Specifically, the intra-chunk vision encoder and inter-chunk feature aggregator execute directly on the GPU, while the language model component leverages vLLM.
This separation can effectively balance CPU-bound preprocessing, compute-intensive visual encoding (Intra/Inter), and language model inference.
More details of the inference efficiency can be found in Appendix B.

\noindent\textbf{Baseline Models}.
We select several representative video models for performance comparison.
We include GPT-4o-20240806~\cite{hurst2024gpt} and Gemini-1.5-Pro-002~\cite{gemini} as the closed-source APIs baselines.
Standard auto-regressive models using resolution-preserving frame sampling are represented by LLaVA-OneVision~\cite{li2024llava} and InternVL 2.5~\cite{chen2024expanding}.
For video understanding tasks, we add models based on:
(a) high-performing sparse frame sampling (NVILA~\cite{liu2024nvila}, LLaVA-Video~\cite{zhang2024video});
(b) dense sampling with lower resolution (Qwen2.5-VL~\cite{qwen2.5vl}, Oryx-1.5~\cite{liu2024oryx});
(c) dense sampling with token compression (VideoChat-Flash~\cite{videochat-flash}, VideoLLaMA3~\cite{zhang2025videollama3});
and (d) slow-fast architecture, a special frame sampling strategy (Slow-fast MLLM~\cite{shi2025slow}).
Regarding image tasks, as some video-centric models either lack image input (e.g., VideoChat-Flash) or are not SOTA on image tasks, we include four strong models on QA/Caption benchmarks: GLM-4V~\cite{wang2024cogvlm}, Qwen2.5-VL, DeepSeek-VL2~\cite{deepseek_vl2} and CogVLM2~\cite{hong2024cogvlm2}.
Crucially, aside from prompt modifications, no benchmark-specific hyperparameters (e.g., frame sampling, resolution) were tuned during evaluation for any model, including Mavors.

\noindent\textbf{Benchmarks}.
Video understanding capabilities are assessed across general knowledge QA (MMWorld~\cite{hemmworld}, PerceptionTest~\cite{patraucean2023perception}), long-video QA (Video-MME~\cite{videomme}, MLVU~\cite{MLVU}), event understanding QA (MVBench~\cite{li2024mvbench}, EventHallusion~\cite{zhang2024eventhallusion}), temporal understanding QA (TempCompass~\cite{tempcompass}, VinoGround~\cite{vinoground}), and captioning (DREAM-1K~\cite{wang2024tarsier}). 
Image understanding evaluation includes comprehensive capabilities (MMMU~\cite{yue2023mmmu}), cognitive understanding (MathVista~\cite{mathvista}, AI2D~\cite{kembhavi2016diagram}), and captioning (CapsBench~\cite{liu2024playground}).
More experiment details can be found in Appendix C.

\subsection{Main Results}

\noindent\textbf{Video Understanding}.
Table~\ref{tab:video} presents a performance comparison of Mavors against baseline models on various video benchmarks.
Approaches employing dense frame sampling with lower resolution demonstrate strong performance on long video QA by incorporating extensive temporal information, but exhibit limitations in understanding spatial details for knowledge-intensive and captioning tasks. 
token compression strategies show a similar pattern, yielding excellent scores on long video QA due to abundant temporal cues, but their merging of non-primary tokens compromises the comprehension of environmental context, resulting in marked deficiencies, especially in captioning.
In contrast, sparse frame sampling approaches, which inherently lose temporal detail and consequently perform less effectively on event understanding QA.
Mavors's multi-granularity video understanding framework successfully balances these trade-offs. 
Leveraging efficient visual information compression, Mavors delivers performance on long video QA nearly on par with dense sampling and token compression techniques, while preserving robust capabilities for knowledge-based and temporal reasoning tasks, eliminating the need for dataset-specific hyperparameter tuning. 
The substantial gains observed for Mavors in captioning highlight the effectiveness in achieving accurate and comprehensive understanding of entire video events.

\noindent\textbf{Image Understanding}.
Table~\ref{tab:image} compares Mavors's performance against baseline models on image benchmarks.
Mavors achieves performance on par with similarly-sized image understanding models in Image QA.
Its captioning performance is particularly strong, surpassing even 72B models.
This effectiveness is partly due to Mavors's architecture: images and videos offer complementary visual perception within the intra-chunk vision encoder, yet are processed without mutual interference by the inter-chunk feature aggregator.

\begin{table}[t]
    \centering
    \resizebox{\linewidth}{!}{
    \begin{tabular}{lccccc}
    \toprule
        \textbf{Model} & \textbf{Size} & \textbf{MMMU} & \textbf{MathVista} & \textbf{AI2D} & \textbf{CapsBench}\\
        \midrule
        \textcolor{lightgray}{GPT-4o-20240806} & \textcolor{lightgray}{-} & \textcolor{lightgray}{69.9} & \textcolor{lightgray}{62.9} & \textcolor{lightgray}{84.7} & \textcolor{lightgray}{67.3}\\
          \textcolor{lightgray}{Gemini-1.5-Pro} & \textcolor{lightgray}{-} & \textcolor{lightgray}{60.6} & \textcolor{lightgray}{58.3} & \textcolor{lightgray}{79.1} & \textcolor{lightgray}{71.2}\\   
        \midrule
        CogVLM2 & 8B & 42.6 & 38.7 & 73.4 & 50.9 \\
        GLM-4V & 9B & 46.9 & 52.2 & 71.2 & 61.0 \\
        LLaVA-OneVision & 7B & 47.9 & 62.6 & 82.4 & 57.4 \\
        InternVL 2.5 & 8B & 56.2 & 64.5 & \textbf{84.6} & 66.5 \\
        Qwen2.5-VL & 7B & \textbf{58.0} & 68.1 & 84.3  & 64.9 \\
        \midrule
        \textcolor{lightgray}{DeepSeek-VL2} & \textcolor{lightgray}{27B} & \textcolor{lightgray}{54.0} & \textcolor{lightgray}{63.9} & \textcolor{lightgray}{83.8} & \textcolor{lightgray}{61.3}\\
        \textcolor{lightgray}{Qwen2.5-VL} & \textcolor{lightgray}{72B} & \textcolor{lightgray}{68.2} & \textcolor{lightgray}{74.2} & \textcolor{lightgray}{88.5} & \textcolor{lightgray}{70.1}\\
        \textcolor{lightgray}{InternVL 2.5} & \textcolor{lightgray}{78B} & \textcolor{lightgray}{70.0} & \textcolor{lightgray}{70.6} & \textcolor{lightgray}{89.1} & \textcolor{lightgray}{68.5}\\
        \midrule
        Mavors (Ours) & 7B & 53.2 & \textbf{69.2} & 84.3 & \textbf{75.2}\\
        \bottomrule
    \end{tabular}
    }
    \caption{Performance on image benchmarks.}
    \vspace{-4mm}
    \label{tab:image}
\end{table}

\begin{figure*}[t]
\centering
\begin{minipage}[t]{0.32\linewidth}
\centering
\includegraphics[width=\linewidth]{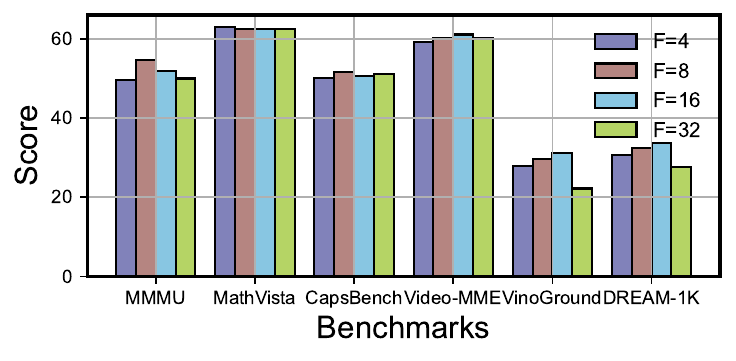}
\caption{Performance with different numbers of frames in a video chunk.}
\label{fig:encoding_frames}
\end{minipage}
\hfill
\begin{minipage}[t]{0.32\linewidth}
\centering
\includegraphics[width=\linewidth]{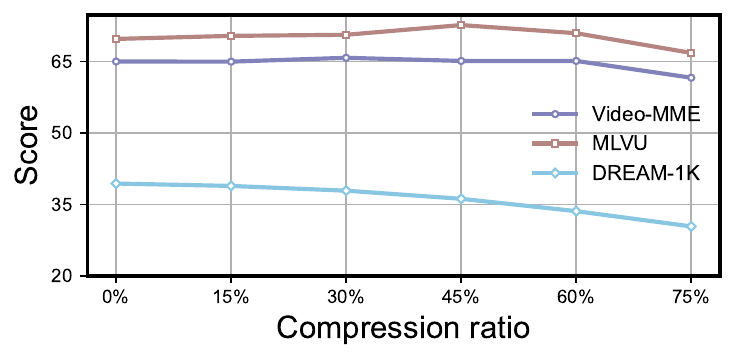}
\caption{Performance with different token compression ratios.}
\label{fig:drop_ratio}
\end{minipage}
\hfill
\begin{minipage}[t]{0.32\linewidth}
\centering
\includegraphics[width=\linewidth]{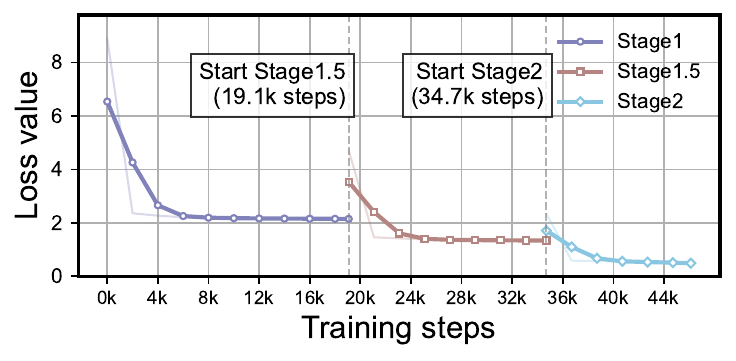}
\caption{The dynamic of training losses across different stages for Mavors.}
\label{fig:loss}
\end{minipage}
\end{figure*}

\subsection{Ablation Studies}

We conduct a series of ablation studies to validate our model design.
Given the extensive training time required for the full training paradigm, these ablations utilize standard compositive datasets and train various versions up to the completion of Stage 2. 
Specifically, Stage 1 employs LLaVA-Pretrain-558K~\cite{liu2023llava15} and LLaVA-Hound-Pretrain~\cite{llava-hound};
Stage 1.5 uses M4-Instruct~\cite{li2024llavanextinterleave} and ShareGPT4o~\cite{cui2025comprehensive};
and Stage 2 utilizes LLaVA-OneVision and LLaVA-Video.
This approach reduces the duration of a full training cycle to under 24 hours with 64 GPUs.
Performance is subsequently monitored using MMMU, MathVista, and CapsBench for image understanding capabilities, and Video-MME, Vinoground, and DREAM-1K for video understanding capabilities.

\noindent\textbf{Effect of the Number of Frames in a Video Chunk}.
We conduct experiments with four settings, varying a parameter $F$ with values of 4, 8, 16, and 32. 
Upon the preliminary study evaluating video captioning performance on the validation set of KVQ~\cite{lu2024kvq}, we observe that configurations with $F=8$ or $F=16$ yield more accurate and comprehensive captions.
To ensure exposure to richer visual information, we finalize the $F=16$ setting.
We further evaluate these four model variants on six benchmark datasets in Figure~\ref{fig:encoding_frames}.
On image-based tasks, we observe a marginal improvement in performance metrics with increasing $F$.
We hypothesize that this improvement stems from the model's increased exposure to individual frames during video processing when $F$ is larger, thereby enhancing its image understanding capabilities.
Conversely, for video understanding tasks, performance degrades significantly for $F=4$ due to insufficient temporal information and for $F=32$, likely due to excessive information compression.


\begin{table}
    \centering
    \resizebox{\columnwidth}{!}{
    \begin{tabular}{lcccccc}
    \toprule
        \textbf{$L_\text{Inter}$} & \textbf{MMMU} & \textbf{MathVista} & \textbf{CapsBench} & \textbf{Video-MME} & \textbf{VinoGround} & \textbf{DREAM-1K} \\
        \midrule
        0 & 50.3 & 63.0 & 51.4 & 61.0 & 27.9 & 30.2 \\
        1 & 51.5 & 63.3 & 50.6 & 60.9 & 30.6 & 32.4 \\
        3 & 52.0 & 62.6 & 50.6 & 61.1 & 31.1 & 33.8 \\ 
        5 & 49.8 & 61.9 & 50.3 & 61.1 & 31.2 & 33.6 \\ 
        \bottomrule
    \end{tabular}
    }
    \caption{Ablation on layers of Transformers in IFA.}
    \label{tab:ifa}
\end{table}

\noindent\textbf{Effect of the IFA Module}.
We establish two baseline models for comparison in Table~\ref{tab:ifa}.
The first baseline completely removes the inter-chunk feature aggregator ($L_\text{Inter}$=0),
where the output from the IVE module is passed directly through a projector and then concatenated with the LLM's input sequence.
In this setup, the integration of temporal and spatial information relies solely on the LLM.
The second baseline utilizes only a single Transformer layer ($L_\text{Inter}$=1) for the aggregator, thereby reducing its computational complexity.
In Table~\ref{tab:ifa},
on image evaluation tasks, removing the Transformer ($L_\text{Inter}$=0) shows a slight advantage, potentially due to the lower parameter count facilitating faster convergence on static perception tasks.
However, for video evaluation, we observe that a deeper inter-chunk feature aggregator ($L_\text{Inter}$=3) enhances the model's understanding, leading to better scores, although with diminishing marginal returns.
Considering model complexity and convergence difficulty, $L_\text{Inter}$=3 should be an efficient configuration of Mavors.

\begin{table}
    \centering
    \resizebox{\columnwidth}{!}{
    \begin{tabular}{lcccccc}
    \toprule
        \textbf{RoPE} & \textbf{MMMU} & \textbf{MathVista} & \textbf{CapsBench} & \textbf{Video-MME} & \textbf{VinoGround} & \textbf{DREAM-1K} \\
        \midrule
        Standard & 51.9 & 62.6 & 50.7 & 61.0 & 30.3 & 32.9 \\ 
        C-RoPE & 52.0 & 62.6 & 50.6 & 61.1 & 31.1 & 33.8 \\ 
        & (+0.1) & (+0.0) & (-0.1) &  (+0.1) & (+0.8) & (+0.9) \\
        \bottomrule
    \end{tabular}
    }
    \caption{Ablation on C-RoPE.}
    \label{tab:crope}
\end{table}

\noindent\textbf{Effect of C-RoPE}.
To assess the performance of C-RoPE, we replace it with the standard RoPE implementation and monitor changes in the Mavors model's visual understanding performance. 
Table~\ref{tab:crope} shows the performance across six metrics. 
For image understanding, given that the IFA architecture processes sub-images independently, both RoPE variants perform comparably.
Conversely, for video understanding, C-RoPE outperforms standard RoPE by an average of 0.6 points. 
It indicates that standard RoPE suffers from differentiating intra-chunk from inter-chunk tokens and may hinder temporal sequence modeling. 
These findings demonstrate the efficacy and importance of C-RoPE within the IFA architecture.

\subsection{Further Analysis}

\noindent\textbf{Analysis on the Ratios of Token Compression}.
We apply token compression techniques within Mavors to decrease the number of tokens on each video chunk.
Specifically, prior to the inter-chunk feature aggregator, we compute similarity between features at corresponding indices in adjacent chunks. 
Tokens exceeding a predefined similarity threshold are merged via averaging, retaining the positional ID from the earlier chunk.
We vary thresholds to achieve different token reduction ratios, summarized in Figure~\ref{fig:drop_ratio}.
Results indicate that Mavors' performance on video QA remains largely unaffected with token reductions up to 60\%.
Conversely, a significant performance degradation is observed for video captioning.
This suggests that token compression on Mavors can be a feasible strategy for reducing inference costs in long-video QA applications.
We provide two representative cases in Appendix F.

\begin{table}[htbp]
  \centering
  \resizebox{\columnwidth}{!}{
    \begin{tabular}{lcccc}
    \toprule
    \textbf{Stage} & \textbf{MMMU} & \textbf{CapsBench} & \textbf{Video-MME} & \textbf{DREAM-1K} \\
    \midrule
    Stage 1  & 36.3 & 54.8 & 48.4 & 23.6 \\
    Stage 1.5 & 47.3 & 62.5 & 53.9 & 26.3 \\
    Stage 2 & 53.0 & 73.4 & 65.0 & 38.9 \\
    DPO & 53.2 & 75.2 & 65.0 & 39.2 \\
    \bottomrule
    \end{tabular}%
  }
  \caption{Results of different training stages.}
  \label{tab:training_stages}%
\end{table}%

\noindent\textbf{Analysis on the Training Dynamics.}
Table~\ref{tab:training_stages} shows the results on the image QA dataset (MMMU), image caption dataset (CapsBench), video QA dataset (Video-MME) and video caption dataset (DREAM-1K) at different stages. The performance on all four datasets improves consistently across the three training stages (Stage 1, Stage 1.5, and Stage 2), indicating that each stage contributes positively to the model's ability to handle different tasks and modalities. The DPO stage provides further improvements.
Note that we also provide the training loss curve of Mavors in Figure~\ref{fig:loss}.

\begin{figure}[ht]
  \centering
  \includegraphics[width=0.9\linewidth]{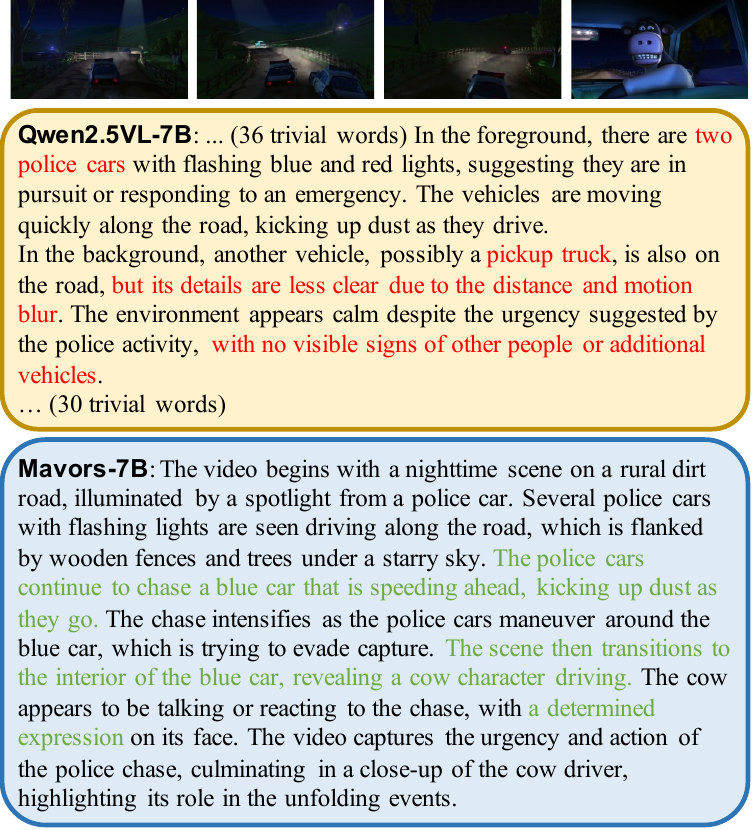}
  \caption{Comparison of  generated video captions from Qwen2.5-VL-7B and Mavors-7B.}\label{fig:showcase}
\end{figure}

\noindent\textbf{Visualization}.
We pick a complex video cut from DREAM-1K and present the captions generated by Qwen2.5VL-7B and Mavors-7B in Figure~\ref{fig:showcase}. 
Despite processing densely sampled frames, Qwen2.5VL-7B fails to capture many details (e.g., omitting the mention of a cow driving), leading to flawed inferences (words in red).
In contrast, we observe that Mavors-7B predicts fine-grained and correct details (words in greed),
which show the effect of our  Mavors-7B.

%% file: sec/6_conclusion.tex
\section{Conclusion}
\label{sec:conclusion}

In this work, we present Mavors, a novel framework for holistic long-context video understanding in MLLMs. Mavors introduces multi-granularity video representation based on Intra-chunk Vision Encoder (IVE) and Inter-chunk Feature Aggregator (IFA) to preserve both spatial details and temporal dynamics and maintain high efficiency.
Extensive experiments on multiple benchmarks demonstrate the effectiveness and efficiency of our Mavors. 

%% file: sec/X_suppl.tex
\clearpage
\setcounter{page}{1}
\setcounter{section}{0} 
\renewcommand{\thesection}{\Alph{section}} 
\maketitlesupplementary

\begin{table}[ht]
    \resizebox{\linewidth}{!}{
    \centering
    \begin{tabular}{|lp{\linewidth}|}
    \hline
    \multicolumn{1}{|l|}{Task} & Dataset \\ \hline
    \multicolumn{2}{|l|}{Stage 1 Datasets} \\ \hline
    \multicolumn{1}{|l|}{Image Caption} & LAION (EN 6.7M, ZH 3.2M) \cite{laion400m}, Conceptual Captions (7.3M) \cite{sharma2018conceptual}, SBU (0.8M) \cite{ordonez2011im2text}, COYO (11M) \cite{coyo-700m}, WuKong (2.9M) \cite{gu2022wukong}, LAION COCO (16M) \cite{lacion-coco}, OMEGA Image Caption (79M) \cite{omegaproject} \\ \hline
    \multicolumn{1}{|l|}{Video Caption} & InternVid-10M-FLT (1.6M) \cite{wang2023internvid}, Panda-70M (0.9M) \cite{chen2024panda}, OMEGA Video Caption (4M) \cite{omegaproject} \\ \hline
    \multicolumn{2}{|l|}{Stage 1.5 Datasets} \\ \hline
    \multicolumn{1}{|l|}{Image Caption} & Met-meme \cite{xu2022metmeme}, PD12M \cite{meyer2024public}, dalle3 \cite{dalle3}, GBC10M \cite{hsieh2024graph}, DenseFusion-1M \cite{li2024densefusion}, GameBunny \cite{taesiri2024videogamebunny}, MERMAID \cite{toh2023mermaid}, CC12M (1M) \cite{changpinyo2021cc12m}, BLIP3 \cite{awadalla2024blip3}, AllSeeingV2 \cite{wang2024all} \\ \hline
    \multicolumn{1}{|l|}{Video Caption} & ChronoMagic \cite{yuan2024chronomagic}, VideoChatGPT \cite{Maaz2023VideoChatGPT}, YouCook2 \cite{zhou2018towards}, CelebV \cite{yu2023celebv}, SthSthV2 \cite{goyal2017something},  MiraData \cite{ju2024miradata}, Hacs \cite{zhao2019hacs}, OpenVid-1M \cite{nan2024openvid}, Kinetics\_700 \cite{carreira2019short}, ShareGPT4Video \cite{chen2024sharegpt4video}, Vript \cite{yang2024vript}, Shot2Story \cite{han2023shot2story20k}, ShareGemini \cite{sharegemini} \\ \hline
    \multicolumn{1}{|l|}{Question Answering} & MMDU \cite{liu2024mmdu}, MMiT \cite{monfort2021multi} \\ \hline
    \multicolumn{1}{|l|}{Knowledge} & Wikipedia \cite{wikidump}, Wikimedia \cite{wikidump}, WIT \cite{srinivasan2021wit} \\ \hline
    \multicolumn{1}{|l|}{Code} & WebSight \cite{laurençon2024unlocking} \\ \hline
    \multicolumn{1}{|l|}{OCR} & LSVT \cite{sun2019icdar}, ArT \cite{chng2019icdar2019}, DocMatix \cite{laurenccon2024building} \\ \hline
    \multicolumn{1}{|l|}{Interleaved} & OBELICS \cite{laurencon2023obelics}, PIN \cite{wang2024pin} \\ \hline
    \multicolumn{1}{|l|}{Mixed-Task Dataset} & MMInstruct \cite{liu2024mminstruct}, LVD-2M \cite{xiong2024lvd}, MMEvol \cite{luo2024mmevol} \\ \hline
    \multicolumn{2}{|l|}{Stage 2 Datasets} \\ \hline
    \multicolumn{1}{|l|}{Instruction} & Countix \cite{Dwibedi_2020_CVPR}, VideoChat \cite{2023videochat}, Videogpt+ \cite{maaz2024videogpt+}, Openmathinstruct-2 (2M) \cite{toshniwal2024openmathinstruct}, RepCountA \cite{hu2022transrac}, Vidgen-1m \cite{tan2024vidgen}, CompCap \cite{chen2024compcap}, Metamath \cite{yu2023metamath}, Llava-Onevision \cite{li2024llava}, Anytext (0.3M) \cite{tuo2023anytext}, Llava-Video \cite{zhang2024video}, S-MiT \cite{monfort2021spoken}, LSMDC \cite{rohrbach2017movie}, Infinity-MM \cite{gu2024infinity}, Mantis \cite{Jiang2024MANTISIM}, ShareGPT4V \cite{chen2023sharegpt4v}, CinePile \cite{rawal2024cinepile}, LLaVA-Hound \cite{llava-hound} \\ \hline
    \multicolumn{1}{|l|}{Grounding} & GRIT \cite{Kosmos2}, RefCOCO \cite{kazemzadeh2014referitgame} \\ \hline
    \multicolumn{1}{|l|}{Temporal Grounding} & GroundedVideoLLM \cite{wang2024grounded} \\ 
    \hline
    \multicolumn{2}{|l|}{Stage 3 (DPO) Datasets} \\ \hline
    \multicolumn{1}{|l|}{Open-ended QA} & Llava-Video \cite{zhang2024video} (10K) \\ \hline
    \multicolumn{1}{|l|}{Image Caption} & Llava-Onevision \cite{li2024llava} (10K), DenseFusion-1M \cite{li2024densefusion} (10K) \\ \hline
    \multicolumn{1}{|l|}{Video Caption} & WebVid \cite{bain2021frozen} (8K), Kinetics\_700 \cite{carreira2019short} (8K), OOPS \cite{epstein2019oops} (4K)\\ \hline
    \end{tabular}}
    \caption{Summary of the training datasets of different stages.}\label{pre-training-datasets}
    
\end{table}

\section{Training Datasets}

The datasets used for training our model at different stages are shown in Table \ref{pre-training-datasets}. For a number of large-scale datasets, we have randomly selected a specific number of samples. The count of these samples is also indicated in Table \ref{pre-training-datasets}. 

We have also curated two datasets from the OMEGA project \cite{omegaproject}, the OMEGA Image Caption (containing 79M samples) and OMEGA Video Caption (containing 4M samples), by sampling videos and images along with their corresponding titles and captions. These two datasets are utilized in the first stage of our model training. 

For certain datasets that either lack captions or only possess low-quality ones, for example, CC12M \cite{changpinyo2021cc12m}, CelebV \cite{yu2023celebv}, Hacs \cite{zhao2019hacs}, and Kinetics\_700 \cite{carreira2019short}, we carefully designed a pipeline to generate high-quality captions. Initially, we utilized Qwen2VL-72B \cite{Qwen2VL}, InternVL2.5-78B-MPO \cite{chen2024expanding} and Tarsier-34B \cite{wang2024tarsier} (video only) to describe these samples in detail. Subsequently, we used DeepSeek-R1-Distill-Llama-70B \cite{deepseekai2025deepseekr1} to amalgamate captions generated by different models while attempting to resolve all inconsistencies using its COT capabilities. The captions produced by this process generally demonstrated superior quality and comprehensibility.

We observed that many composite datasets incorporate content from established standalone datasets, leading to potential data redundancy.
To address this, we implemented a deduplication process for identical samples (images or videos).
Specifically, we calculated the Perplexity (PPL) of the associated text using the Qwen2VL-72B \cite{Qwen2VL} model, distinguishing between QA and Captioning tasks. 
For duplicate visual content within QA tasks, we retained the two samples exhibiting the lowest text PPL scores.
For Captioning tasks, one sample was randomly selected from the two with the lowest PPL for inclusion in our training set.

For the data in the DPO stage, we selected a specific number of samples from the corresponding datasets. The preference datasets were then generated in accordance with the following methods:

\begin{enumerate}
\item Open-ended QA: Positive examples are generated by prompting the model with diverse inputs to produce responses that are correct, of appropriate length, and properly terminated. Negative examples are derived from the same inputs by adjusting the sampling temperature to elicit incorrect or overly brief answers.
\item Image Captioning: Multiple candidate captions are generated per image using the model under high temperatures. These candidates are then ranked according to a predefined scoring strategy, forming positive (higher-ranked) and negative (lower-ranked) pairs for DPO training.
\item Video Captioning: Captions generated from the original video serve as positive examples. Negative examples are created by captioning the video after segmenting it into four equal parts and shuffling their temporal order.
\end{enumerate}

\section{Analysis on the Inference Costs}

\begin{table}
    \resizebox{\linewidth}{!}{
    \centering
    \begin{tabular}{llcc}
    \toprule
        & & \textbf{Qwen2.5VL-7B} & \textbf{Mavors-7B}\\
        \midrule
        \multirow{2}{*}{Images} & Prefilling (ms) & 397 & 392\\
        & Decoding (token/s) & 23 & 30\\
        \midrule
        \multirow{2}{*}{Videos} & Prefilling (ms) & 1,225 & 448\\
        & Decoding (token/s) & 22 & 30 \\
        \bottomrule
    \end{tabular}}
    \caption{Inference efficiency between Qwen2.5VL-7B and Mavors-7B. Model is better when Prefilling (ms) is lower and Decoding (token/s) is larger.}
    \label{tab:inference}
\end{table}

We evaluate the inference performance of Qwen2.5VL-7B and Mavors-7B using an GPU. 
Initially, we measure the execution time of the \texttt{model.generate} function via the standard HuggingFace implementation (with FlashAttention-2 enabled) to capture the core model execution time, excluding video preprocessing.
Table~\ref{tab:inference} summarizes the inference times for both models on the DREAM-1K and CapsBench video captioning tasks. 
The results show that Mavors' more efficient video representation reduces both the ViT computations and the language model's context window requirements. 
Consequently, Mavors-7B demonstrates significant speed improvements on video understanding tasks, achieving 2.7x faster prefill and 1.4x faster decoding compared to Qwen2.5VL-7B. 
Furthermore, integrating the vLLM inference framework with overlapping vision preprocessing enables 2.5s per image in CapsBench and 3.7s per video in DREAK-1K, reducing from about 13s per image and 20s per video respectively. 
These findings indicate that Mavors provides an economical solution for scenarios requiring frequent or high-volume multimodal model inference.

\section{Details of Experiments}
\noindent\textbf{Evaluation Setup.}
To ensure a standardized and reproducible evaluation, we conduct experiments on both open-source and closed-source models using consistent protocols. For open-source models, we adopt the lmms-eval framework~\cite{zhang2024lmms}, which offers a unified pipeline tailored for benchmarking MLLMs. All open-source models are evaluated using the officially released checkpoints to preserve the integrity of reported results.
To maintain experimental stability, we fix the decoding strategy to greedy decoding, set the maximum number of generated tokens to 1024. Image and video resolution, along with other preprocessing settings, follow the default configurations provided by the lmms-evak framework or the respective model implementations.
For closed-source models, including Gemini-1.5-Pro-002~\cite{gemini} and GPT-4o-20240806~\cite{hurst2024gpt}, we access them through their official APIs. Due to the restricted controllability over decoding parameters, we adopt the default generation settings provided by each platform.
For benchmarks requiring GPT-based automatic scoring, such as those involving instruction-following or open-ended generation tasks, we follow the evaluation protocol described in the original benchmark papers or apply the default settings specified by the lmms-eval framework to select the judge model. 
Specifically, for MathVista~\cite{mathvista}, we use GPT-4-Turbo (1106) as the judge model. For CapsBench~\cite{liu2024playground} and MMMU~\cite{yue2023mmmu}, we adopt GPT-4o (20240806), while for DREAM-1K~\cite{wang2024tarsier}, we follow the original benchmark and employ GPT-3.5-Turbo (0125) to perform automatic scoring. These choices align with the evaluation protocols used in the respective benchmark papers, ensuring fair and comparable results across models.

\noindent\textbf{Baseline Models.}
To comprehensively evaluate the performance of our proposed Mavors-7B, we select a diverse set of baseline models tailored to the specific characteristics of both image and video benchmarks. 

For image benchmarks, we compare against two leading proprietary models, GPT-4o~\cite{hurst2024gpt} and Gemini-1.5-Pro~\cite{gemini}. GPT-4o, developed by OpenAI, is capable of processing text, images, and audio in a unified manner and has demonstrated strong performance in visual reasoning tasks. Gemini, developed by Google DeepMind, similarly integrates multimodal capabilities and excels in scenarios requiring complex cross-modal understanding. We also include a range of high-performing open-source MLLMs in our comparison. These include CogVLM2~\cite{hong2024cogvlm2}, a model optimized for visual-language understanding in dynamic contexts; GLM-4V~\cite{hong2024cogvlm2}, which extends the GLM architecture with strong visual recognition capabilities; LLaVA-OneVision~\cite{li2024llava}, a widely recognized open-source MLLM that integrates a collection of high-quality multimodal datasets, advanced training strategies, and refined model designs to achieve strong performance across image-based benchmarks; InternVL2.5~\cite{chen2024expanding}, which is an advanced MLLM series developed by Shanghai Artificial Intelligence Laboratory. Building upon the architecture of InternVL2, it introduces significant enhancements in training strategies and data quality; DeepSeek-VL2~\cite{deepseek_vl2}, an MoE-based model balancing scalability and accuracy; and Qwen2.5-VL~\cite{qwen2.5vl}, a model that significantly enhance general image recognition capabilities, expanding to a vast array of categories, including plants, animals, landmarks, and various products. It also excels in precise object localization, advanced text recognition, and document parsing.

For video benchmarks, we select four representative categories of baseline models, each exemplifying distinct video processing strategies. The first category includes models that employ sparse frame sampling with high performance, such as NVILA~\cite{liu2024nvila} and LLaVA-Video~\cite{llava-video-sft}, which focus on selecting key frames to reduce computational overhead while maintaining contextual understanding. NVILA, developed by NVIDIA, utilizes a “scale-then-compress” paradigm that first increases spatial and temporal resolutions and then compresses visual tokens, enabling efficient processing of high-resolution images and long videos. LLaVA-Video improves video understanding by introducing a high-quality synthetic dataset, LLaVA-Video-178K~\cite{llava-video-sft}, specifically designed for video instruction-following tasks. Models like Qwen2.5-VL~\cite{qwen2.5vl} and Oryx-1.5~\cite{liu2024oryx} adopt dense frame sampling at lower resolutions to achieve a trade-off between information richness and efficiency (we set at most 768 frames in our experiments). Oryx-1.5 is a unified MLLM designed to flexibly and efficiently handle visual inputs with varying spatial scales and temporal lengths, making it well-suited for processing both high-resolution images and extended video sequences. In addition, we include models such as VideoChat-Flash~\cite{videochat-flash} and VideoLLaMA3~\cite{zhang2025videollama3}, which apply dense sampling combined with token compression to handle long video sequences effectively (up to 1000 frames in our experiments). VideoChat-Flash leverages this strategy to mitigate the computational overhead introduced by dense sampling, enabling effective handling of long-duration inputs without sacrificing performance. Similarly, VideoLLaMA3 integrates token compression with dense sampling to reduce input redundancy, thereby enhancing the model’s ability to understand and reason over extended video content. Finally, we include Slow-fast MLLM~\cite{shi2025slow}, which employs a specialized dual-pathway sampling mechanism to capture temporal dynamics at multiple granularities. By processing visual inputs through both slow and fast pathways, the model effectively models temporal variations across different timescales.

\noindent\textbf{Benchmarks.}
It is crucial to comprehensively and objectively assess a model’s capabilities across various aspects and dimensions. To this end, we include a broad range of representative image and video benchmarks in our evaluation.

We adopt MMMU~\cite{yue2023mmmu}, MathVista~\cite{mathvista}, AI2D~\cite{kembhavi2016diagram}, and CapsBench~\cite{liu2024playground} as representative image benchmarks, covering a broad range of visual understanding and reasoning tasks. 
\begin{itemize}
\item \textbf{MMMU} targets expert-level multimodal reasoning across diverse academic domains, featuring varied visual inputs such as charts, diagrams, and tables. 
\item \textbf{MathVista} focuses on complex mathematical problem solving that integrates textual and visual information. 
\item \textbf{AI2D} evaluates the ability to interpret scientific diagrams commonly used in elementary science education. 
\item \textbf{CapsBench} emphasizes compositional reasoning by requiring models to generate comprehensive, detailed, and accurate descriptions of visual scenes. It challenges models to precisely capture a wide range of visual information, including object attributes, spatial relationships, and inter-object interactions. 
\end{itemize}
Together, these benchmarks offer a comprehensive assessment of image-based multimodal capabilities.

We conduct evaluations on a diverse set of video benchmarks, including MMWorld~\cite{hemmworld}, PerceptionTest~\cite{patraucean2023perception}, Video-MME~\cite{videomme}, MLVU~\cite{MLVU}, MVBench~\cite{li2024mvbench}, EventHallusion~\cite{zhang2024eventhallusion}, TempCompass~\cite{tempcompass}, VinoGround~\cite{vinoground}, and DREAM-1K~\cite{wang2024tarsier}. 
\begin{itemize}
\item \textbf{MMWorld} evaluates MLLMs’ ability to reason about real-world dynamics across diverse disciplines and tasks. It includes 1,910 videos and 6,627 QA pairs covering explanation, counterfactual reasoning, and future prediction. 
\item \textbf{PerceptionTest} evaluates the perceptual and reasoning skills of MLLMs across video, audio, and text modalities. It includes 11.6K real-world videos and focuses on cognitive skills and reasoning types—such as memory, abstraction, and counterfactual thinking—beyond traditional classification or detection tasks. We use the validation set in the experiments.
\item \textbf{Video-MME} is a comprehensive benchmark for evaluating MLLMs across diverse video types, temporal lengths, and multimodal inputs including subtitles and audio. It features 900 manually annotated videos spanning 254 hours and 2,700 QA pairs, offering a rigorous test of models’ generalization and contextual understanding. We evaluate Video-MME without subtitles in our experiments. 
\item \textbf{MLVU} is a benchmark designed for comprehensive evaluation of long video understanding, featuring extended video durations and diverse genres such as movies, surveillance, and egocentric videos. It includes a variety of tasks to assess MLLMs’ abilities in handling complex temporal dependencies and multi-scene reasoning across long-form content. 
\item \textbf{MVBench} is a diagnostic benchmark designed to evaluate the temporal understanding capabilities of MLLMs through 20 challenging video tasks that go beyond static image reasoning. By systematically transforming static tasks into dynamic ones, it covers a wide range of temporal skills and ensures fair evaluation using ground-truth annotations converted into multiple-choice questions. 
\item \textbf{EventHallusion} is a benchmark designed to evaluate hallucination in MLLMs, specifically focusing on event-level understanding—a core aspect of video analysis. It probes models’ susceptibility to language priors and vision-language biases, providing a targeted assessment of their reliability in temporal event reasoning. 
\item \textbf{TempCompass} is a benchmark designed to evaluate the fine-grained temporal perception abilities of MLLMs across diverse task types. By introducing videos with controlled temporal variations and minimizing static or linguistic bias, it enables precise assessment of model performance on aspects such as speed, direction, and sequence understanding. 
\item \textbf{VinoGround} is a benchmark that evaluates temporal counterfactual reasoning in short videos through 1,000 natural video-caption pairs.
\item \textbf{DREAM-1K} is a challenging benchmark for detailed video description, featuring 1,000 clips from diverse sources such as films, stock footage, and short-form videos. Each video is paired with fine-grained human-annotated descriptions, and evaluated using AutoDQ, a metric better suited for assessing rich, multi-event narratives than traditional captioning scores. 
\end{itemize}
These benchmarks collectively cover a wide range of video understanding challenges, such as temporal reasoning, event prediction, visual grounding, perception under uncertainty, and multi-turn video-based instruction following, enabling a comprehensive assessment of the model’s performance across different video-centric tasks.

\section{Needle in a Haystack Test}

\begin{figure}[htp]
  \centering
  \includegraphics[width=1.0\linewidth]{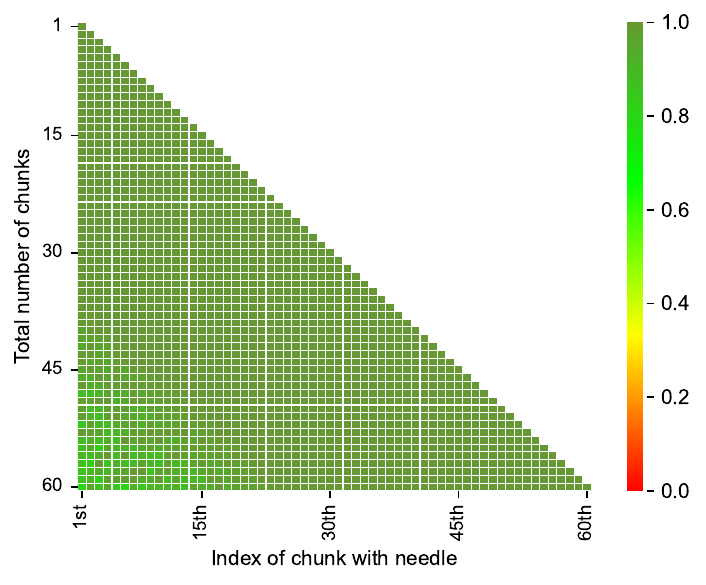}
  \caption{Results of NIAH of Mavors with at most 60 video chunks.}\label{fig:niah}
\end{figure}

Inspired by the design in LongVA~\cite{zhang2024longva}, we conduct Needle-in-a-Haystack (NIAH) test.
We adopt a chunk-level NIAH evaluation scheme, which primarily focuses on evaluating the model's comprehension accuracy when a target frame is inserted into different video chunks.
We utilize 10 short-duration and 10 medium-duration videos from the Video-MME benchmark. 
We examine the model's performance across video lengths ranging from 1 to 60 chunks.
Recall that 60 chunks correspond to 960 frames. 
For a given number of chunks $c_\text{V}$, we performed 50*$c_\text{V}$ trials.
In each trial, we randomly select a video, an image (the `needle'), a specific chunk within the video, and a frame position within that chunk. 
The selected image then replaces the original frame at the chosen position.
Notably, after selecting a video, we first apply accelerating playback (temporal subsampling) to ensure the video frames precisely fit into $c_\text{V}$ chunks. 
Figure~\ref{fig:niah} illustrates the accuracy results. 
As observed, perfect accuracy is achieved within the model's training window length (32 chunks). 
Moreover, Mavors maintains strong accuracy even as the number of chunks increases beyond this window. 
This experiment indicates that Mavors can provide reliable understanding for videos of reasonable duration, provided essential frame information is not lost during the accelerating playback process.

\section{Showcases of Mavors in Image Captioning}
\begin{figure*}[htp]
  \centering
    \includegraphics[width=1.0\linewidth]{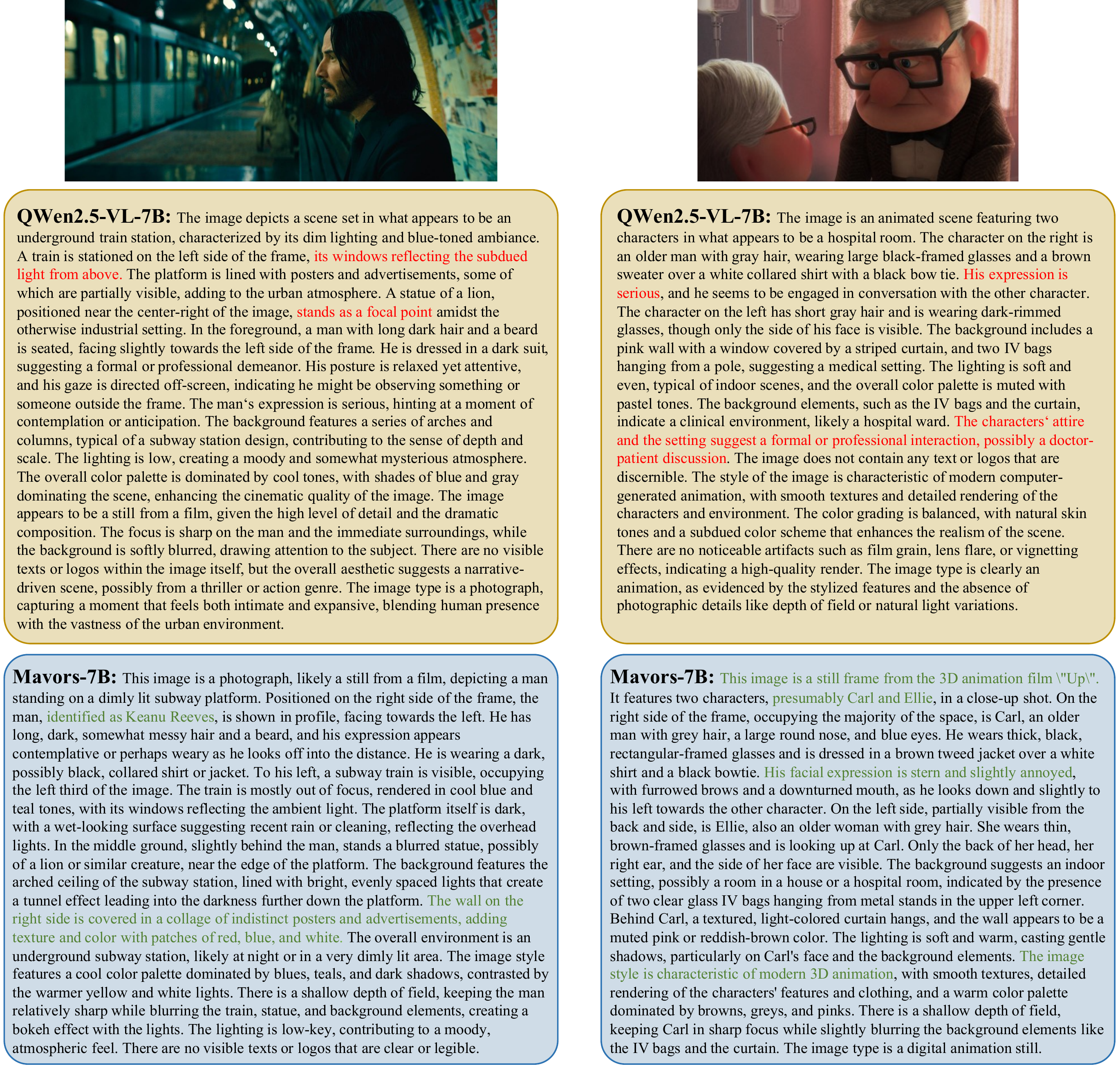}
    \caption{Comparison of the generated image captions from Qwen2.5-VL-7B and Mavors-7B. The text in red contains wrong content, and the text in green marks the detailed descriptions only appear in Mavors.}\label{fig:caps_show}
\end{figure*}

We present a few examples of Mavors' performance on the CapsBench benchmark in Figure~\ref{fig:caps_show}. Mavors demonstrates a strong ability to generate accurate and detailed image captions, and it could recognize more proper nouns and understand human interactions and emotions more precisely than Qwen2.5-VL-7B.

\section{Showcases of Mavors with Token Compression}

\begin{figure*}[htp]
  \centering
  \includegraphics[width=1.0\linewidth]{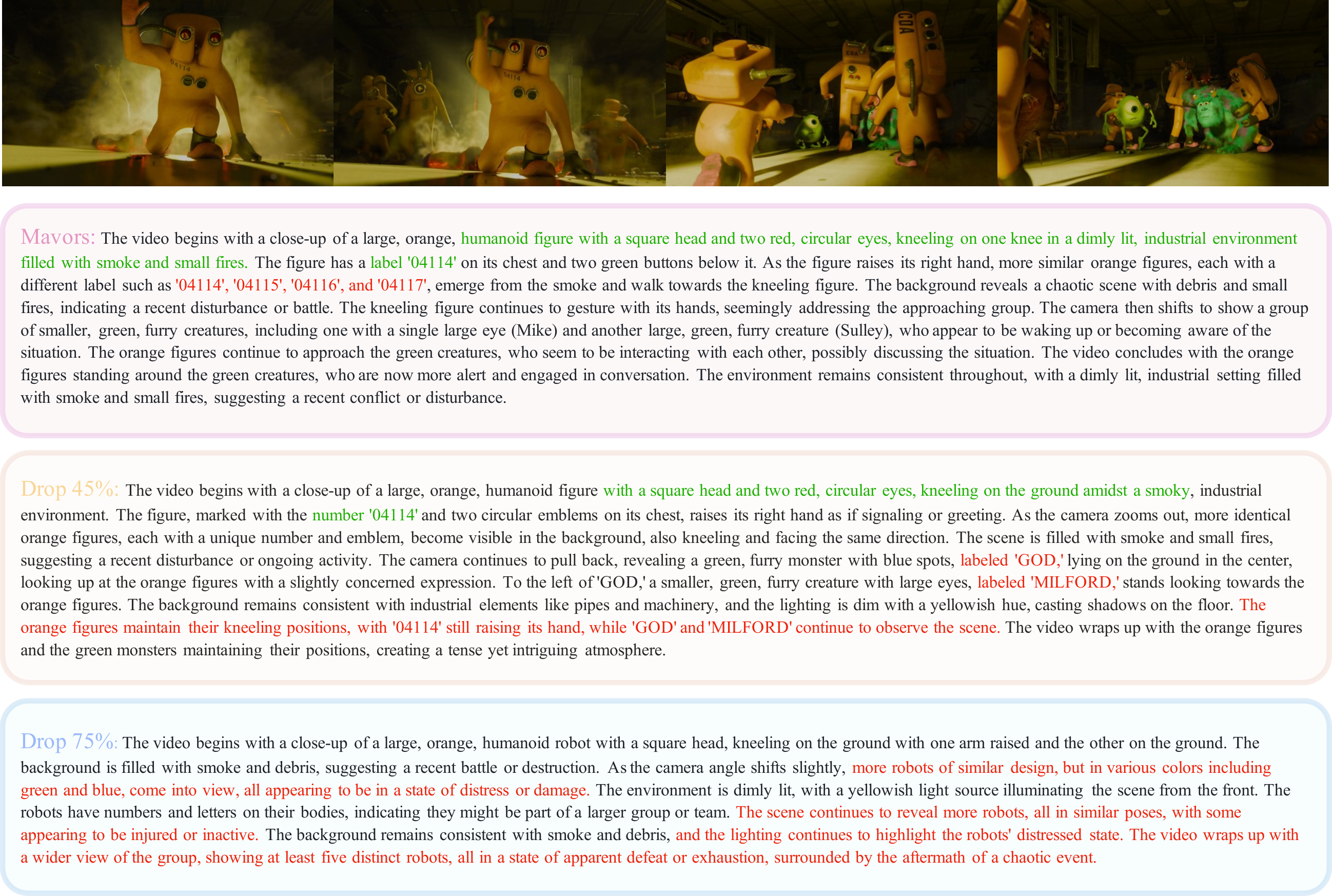}
  \caption{Example of captioning task with token compression: higher compression ratio leads to the missing of critical details.}\label{fig:drop_case_1}
\end{figure*}

\begin{figure*}[htp]
  \centering
  \includegraphics[width=1.0\linewidth]{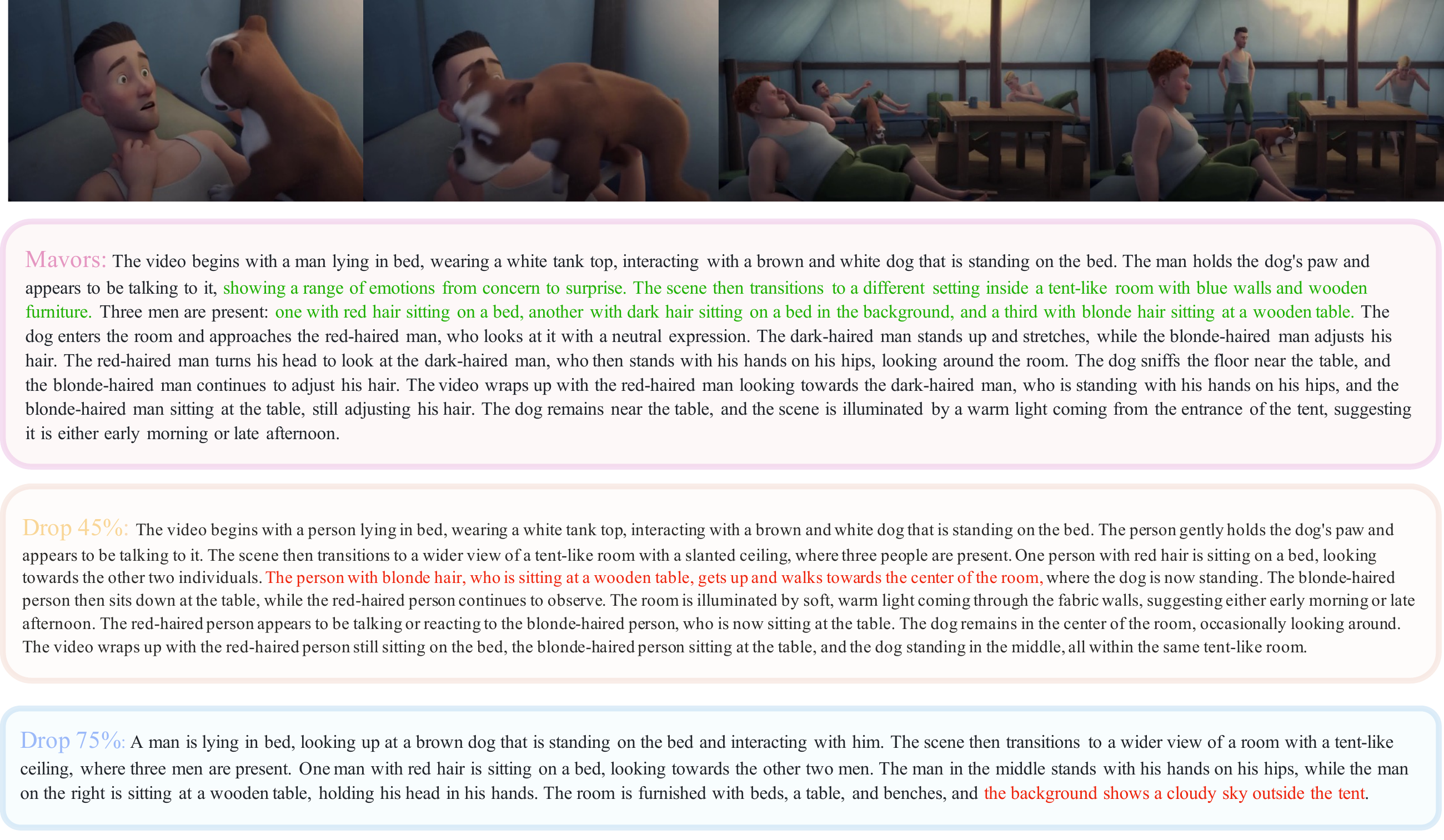}
  \caption{Example of captioning task with token compression: higher compression ratio leads to the missing of critical details.}\label{fig:drop_case_2}
\end{figure*}

Our analysis indicates that as token compression increases up to 60\%, there is negligible impact on Video QA performance, while performance on Captioning tasks degrades progressively. 
We present two case studies to illustrate the specific effects of captioning. 
In the first case (Figure~\ref{fig:drop_case_1}), we observe that despite an imperfect initial caption, higher compression rates lead to increased model hallucinations. 
The model appears to perceive visual changes from the retained tokens but resorts to speculation, resulting in inaccurate descriptions.
In the second case (Figure~\ref{fig:drop_case_2}), increased compression causes the model to generate sparser descriptions, omitting critical details and introducing hallucinations. 
These findings suggest that token compression can pose performance risks, particularly for complex scene captioning tasks.

%% file: main.bib
@misc{yao2024minicpmvgpt4vlevelmllm,
      title={MiniCPM-V: A GPT-4V Level MLLM on Your Phone}, 
      author={Yuan Yao and Tianyu Yu and Ao Zhang and Chongyi Wang and Junbo Cui and Hongji Zhu and Tianchi Cai and Haoyu Li and Weilin Zhao and Zhihui He and Qianyu Chen and Huarong Zhou and Zhensheng Zou and Haoye Zhang and Shengding Hu and Zhi Zheng and Jie Zhou and Jie Cai and Xu Han and Guoyang Zeng and Dahai Li and Zhiyuan Liu and Maosong Sun},
      year={2024},
      eprint={2408.01800},
      archivePrefix={arXiv},
      primaryClass={cs.CV},
      url={https://arxiv.org/abs/2408.01800}, 
}

@inproceedings{siglip,
  title={Sigmoid loss for language image pre-training},
  author={Zhai, Xiaohua and Mustafa, Basil and Kolesnikov, Alexander and Beyer, Lucas},
  booktitle={Proceedings of the IEEE/CVF International Conference on Computer Vision},
  pages={11975--11986},
  year={2023}
}

@article{Qwen2VL,
  title={Qwen2-VL: Enhancing Vision-Language Model's Perception of the World at Any Resolution},
  author={Wang, Peng and Bai, Shuai and Tan, Sinan and Wang, Shijie and Fan, Zhihao and Bai, Jinze and Chen, Keqin and Liu, Xuejing and Wang, Jialin and Ge, Wenbin and Fan, Yang and Dang, Kai and Du, Mengfei and Ren, Xuancheng and Men, Rui and Liu, Dayiheng and Zhou, Chang and Zhou, Jingren and Lin, Junyang},
  journal={arXiv preprint arXiv:2409.12191},
  year={2024}
}

@article{TanKoala2024,
  author = {Reuben Tan and Ximeng Sun and Ping Hu and Jui-hsien Wang and Hanieh Deilamsalehy and Bryan A. Plummer and Bryan Russell and Kate Saenko},
  title = {Koala: Key frame-conditioned long video-LLM},
  year = 2024,
  booktitle={The IEEE Conference on Computer Vision and Pattern Recognition (CVPR)},
}

@article{song2023moviechat,
  title={MovieChat: From Dense Token to Sparse Memory for Long Video Understanding},
  author={Song, Enxin and Chai, Wenhao and Wang, Guanhong and Zhang, Yucheng and Zhou, Haoyang and Wu, Feiyang and Guo, Xun and Ye, Tian and Lu, Yan and Hwang, Jenq-Neng and others},
  journal={arXiv preprint arXiv:2307.16449},
  year={2023}
}

@misc{llava-video-sft,
    title={Video Instruction Tuning With Synthetic Data}, 
    author={Yuanhan Zhang and Jinming Wu and Wei Li and Bo Li and Zejun Ma and Ziwei Liu and Chunyuan Li},
    year={2024},
    eprint={2410.02713},
    archivePrefix={arXiv},
    primaryClass={cs.CV},
    url={https://arxiv.org/abs/2410.02713}, 
}

@misc{liu2024llavanext,
 author = {Liu, Haotian and Li, Chunyuan and Li, Yuheng and Li, Bo and Zhang, Yuanhan and Shen, Sheng and Lee, Yong Jae},
 title = {LLaVA-NeXT: Improved reasoning, OCR, and world knowledge},
 year = {2024}
}

@article{li2024llavanextinterleave,
  title={Llava-next-interleave: Tackling multi-image, video, and 3d in large multimodal models},
  author={Li, Feng and Zhang, Renrui and Zhang, Hao and Zhang, Yuanhan and Li, Bo and Li, Wei and Ma, Zejun and Li, Chunyuan},
  journal={arXiv preprint arXiv:2407.07895},
  year={2024}
}

@inproceedings{yue2023mmmu,
 author = {Xiang Yue and Yuansheng Ni and Kai Zhang and Tianyu Zheng and Ruoqi Liu and Ge Zhang and Samuel Stevens and Dongfu Jiang and Weiming Ren and Yuxuan Sun and Cong Wei and Botao Yu and Ruibin Yuan and Renliang Sun and Ming Yin and Boyuan Zheng and Zhenzhu Yang and Yibo Liu and Wenhao Huang and Huan Sun and Yu Su and Wenhu Chen},
 booktitle = {Proceedings of CVPR},
 title = {MMMU: A Massive Multi-discipline Multimodal Understanding and Reasoning Benchmark for Expert AGI},
 year = {2024}
}

@article{mathvista,
 author = {Lu, Pan and Bansal, Hritik and Xia, Tony and Liu, Jiacheng and Li, Chunyuan and Hajishirzi, Hannaneh and Cheng, Hao and Chang, Kai-Wei and Galley, Michel and Gao, Jianfeng},
 journal = {ArXiv preprint},
 title = {MathVista: Evaluating Math Reasoning in Visual Contexts with GPT-4V, Bard, and Other Large Multimodal Models},
 volume = {abs/2310.02255},
 year = {2023}
}

@article{llava-hound,
 author = {Ruohong Zhang and
Liangke Gui and
Zhiqing Sun and
Yihao Feng and
Keyang Xu and
Yuanhan Zhang and
Di Fu and
Chunyuan Li and
Alexander Hauptmann and
Yonatan Bisk and
Yiming Yang},
 journal = {ArXiv preprint},
 title = {Direct Preference Optimization of Video Large Multimodal Models from
Language Model Reward},
 volume = {abs/2404.01258},
 year = {2024}
}

@misc{liu2023llava15,
 author = {Liu, Haotian and Li, Chunyuan and Li, Yuheng and Lee, Yong Jae},
 journal = {ArXiv preprint},
 title = {Improved Baselines with Visual Instruction Tuning},
 volume = {abs/2310.03744},
 year = {2023}
}

@inproceedings{radford2021clip,
 author = {Alec Radford and
Jong Wook Kim and
Chris Hallacy and
Aditya Ramesh and
Gabriel Goh and
Sandhini Agarwal and
Girish Sastry and
Amanda Askell and
Pamela Mishkin and
Jack Clark and
Gretchen Krueger and
Ilya Sutskever},
 bibsource = {dblp computer science bibliography, https://dblp.org},
 booktitle = {Proceedings of the 38th International Conference on Machine Learning,
{ICML} 2021, 18-24 July 2021, Virtual Event},
 editor = {Marina Meila and
Tong Zhang},
 pages = {8748--8763},
 series = {Proceedings of Machine Learning Research},
 title = {Learning Transferable Visual Models From Natural Language Supervision},
 volume = {139},
 year = {2021}
}

@inproceedings{goyal2017something,
 author = {Raghav Goyal and
Samira Ebrahimi Kahou and
Vincent Michalski and
Joanna Materzynska and
Susanne Westphal and
Heuna Kim and
Valentin Haenel and
Ingo Fr{\"{u}}nd and
Peter Yianilos and
Moritz Mueller{-}Freitag and
Florian Hoppe and
Christian Thurau and
Ingo Bax and
Roland Memisevic},
 bibsource = {dblp computer science bibliography, https://dblp.org},
 booktitle = {{IEEE} International Conference on Computer Vision, {ICCV} 2017, Venice,
Italy, October 22-29, 2017},
 doi = {10.1109/ICCV.2017.622},
 pages = {5843--5851},
 title = {The "Something Something" Video Database for Learning and Evaluating
Visual Common Sense},
 year = {2017}
}

@misc{chen2023sharegpt4v,
 archiveprefix = {arXiv},
 author = {Lin Chen and Jinsong Li and Xiaoyi Dong and Pan Zhang and Conghui He and Jiaqi Wang and Feng Zhao and Dahua Lin},
 eprint = {2311.12793},
 primaryclass = {cs.CV},
 title = {ShareGPT4V: Improving Large Multi-Modal Models with Better Captions},
 year = {2023}
}

@misc{cui2025comprehensive,
            author       = {Erfei Cui and Yinan He and Zheng Ma and Zhe Chen and Hao Tian and Weiyun Wang and Kunchang Li and Yi Wang and Wenhai Wang and Xizhou Zhu and Lewei Lu and Tong Lu and Yali Wang and Limin Wang and Yu Qiao and Jifeng Dai},
            title        = {ShareGPT-4o: Comprehensive Multimodal Annotations With GPT-4o},
            year         = {2024},
            url          = {https://sharegpt4o.github.io/}, 
}

@article{laion400m,
 author = {Christoph Schuhmann and Richard Vencu and Romain Beaumont and Robert Kaczmarczyk and Clayton Mullis and Aarush Katta and Theo Coombes and Jenia Jitsev and Aran Komatsuzaki},
 journal = {ArXiv preprint},
 title = {LAION-400M: Open Dataset of CLIP-Filtered 400 Million Image-Text Pairs},
 volume = {abs/2111.02114},
 year = {2021}
}

@inproceedings{laurencon2023obelics,
 author = {Hugo Lauren{\c{c}}on and
Lucile Saulnier and
L{\'{e}}o Tronchon and
Stas Bekman and
Amanpreet Singh and
Anton Lozhkov and
Thomas Wang and
Siddharth Karamcheti and
Alexander M. Rush and
Douwe Kiela and
Matthieu Cord and
Victor Sanh},
 bibsource = {dblp computer science bibliography, https://dblp.org},
 booktitle = {Advances in Neural Information Processing Systems 36: Annual Conference
on Neural Information Processing Systems 2023, NeurIPS 2023, New Orleans,
LA, USA, December 10 - 16, 2023},
 editor = {Alice Oh and
Tristan Naumann and
Amir Globerson and
Kate Saenko and
Moritz Hardt and
Sergey Levine},
 title = {{OBELICS:} An Open Web-Scale Filtered Dataset of Interleaved Image-Text
Documents},
 year = {2023}
}

@article{vinoground,
          title={Vinoground: Scrutinizing LMMs over Dense Temporal Reasoning with Short Videos},
          author={Zhang, Jianrui and Mu, Cai and Lee, Yong Jae},
          journal={arXiv preprint arXiv:2410.02763},
          year={2024} 
 }

@article{kangaroogroup,
	title={Kangaroo: A Powerful Video-Language Model Supporting Long-context Video Input},
	author={Liu, Jiajun and Wang, Yibing and Ma, Hanghang and Wu, Xiaoping and Ma, Xiaoqi and Wei, xiaoming and Jiao, Jianbin and Wu, Enhua and Hu, Jie},
	journal={arXiv preprint arXiv:2408.15542},
	year={2024}
}

@inproceedings{Jiang2025TokenEfficientLV,
  title={Token-Efficient Long Video Understanding for Multimodal LLMs},
  author={Jindong Jiang and Xiuyu Li and Zhijian Liu and Muyang Li and Guo Chen and Zhiqi Li and De-An Huang and Guilin Liu and Zhiding Yu and Kurt Keutzer and Sungjin Ahn and Jan Kautz and Hongxu Yin and Yao Lu and Song Han and Wonmin Byeon},
  year={2025},
  url={https://api.semanticscholar.org/CorpusID:276813638}
}

@inproceedings{liu2023llava,
 author = {Haotian Liu and
Chunyuan Li and
Qingyang Wu and
Yong Jae Lee},
 bibsource = {dblp computer science bibliography, https://dblp.org},
 booktitle = {Advances in Neural Information Processing Systems 36: Annual Conference
on Neural Information Processing Systems 2023, NeurIPS 2023, New Orleans,
LA, USA, December 10 - 16, 2023},
 editor = {Alice Oh and
Tristan Naumann and
Amir Globerson and
Kate Saenko and
Moritz Hardt and
Sergey Levine},
 title = {Visual Instruction Tuning},
 year = {2023}
}

@article{2023videochat,
 author = {Li, Kunchang and He, Yinan and Wang, Yi and Li, Yizhuo and Wang, Wenhai and Luo, Ping and Wang, Yali and Wang, Limin and Qiao, Yu},
 journal = {ArXiv preprint},
 title = {VideoChat: Chat-Centric Video Understanding},
 volume = {abs/2305.06355},
 year = {2023}
}

@inproceedings{kazemzadeh2014referitgame,
 author = {Kazemzadeh, Sahar  and
Ordonez, Vicente  and
Matten, Mark  and
Berg, Tamara},
 booktitle = {Proceedings of the 2014 Conference on Empirical Methods in Natural Language Processing ({EMNLP})},
 doi = {10.3115/v1/D14-1086},
 editor = {Moschitti, Alessandro  and
Pang, Bo  and
Daelemans, Walter},
 pages = {787--798},
 title = {{R}efer{I}t{G}ame: Referring to Objects in Photographs of Natural Scenes},
 year = {2014}
}

@inproceedings{Alayrac2022FlamingoAV,
 author = {Jean{-}Baptiste Alayrac and
Jeff Donahue and
Pauline Luc and
Antoine Miech and
Iain Barr and
Yana Hasson and
Karel Lenc and
Arthur Mensch and
Katherine Millican and
Malcolm Reynolds and
Roman Ring and
Eliza Rutherford and
Serkan Cabi and
Tengda Han and
Zhitao Gong and
Sina Samangooei and
Marianne Monteiro and
Jacob L. Menick and
Sebastian Borgeaud and
Andy Brock and
Aida Nematzadeh and
Sahand Sharifzadeh and
Mikolaj Binkowski and
Ricardo Barreira and
Oriol Vinyals and
Andrew Zisserman and
Kar{\'{e}}n Simonyan},
 bibsource = {dblp computer science bibliography, https://dblp.org},
 booktitle = {Advances in Neural Information Processing Systems 35: Annual Conference
on Neural Information Processing Systems 2022, NeurIPS 2022, New Orleans,
LA, USA, November 28 - December 9, 2022},
 editor = {Sanmi Koyejo and
S. Mohamed and
A. Agarwal and
Danielle Belgrave and
K. Cho and
A. Oh},
 title = {Flamingo: a Visual Language Model for Few-Shot Learning},
 year = {2022}
}

@inproceedings{changpinyo2021cc12m,
 author = {Soravit Changpinyo and
Piyush Sharma and
Nan Ding and
Radu Soricut},
 bibsource = {dblp computer science bibliography, https://dblp.org},
 booktitle = {{IEEE} Conference on Computer Vision and Pattern Recognition, {CVPR}
2021, virtual, June 19-25, 2021},
 doi = {10.1109/CVPR46437.2021.00356},
 pages = {3558--3568},
 title = {Conceptual 12M: Pushing Web-Scale Image-Text Pre-Training To Recognize
Long-Tail Visual Concepts},
 year = {2021}
}

@article{gemini,
 author = {{Gemini Team}},
 journal = {ArXiv preprint},
 title = {Gemini 1.5: Unlocking multimodal understanding across millions of tokens of context},
 volume = {abs/2403.05530},
 year = {2024}
}

@article{li2024llava,
 author = {Li, Bo and Zhang, Yuanhan and Guo, Dong and Zhang, Renrui and Li, Feng and Zhang, Hao and Zhang, Kaichen and Li, Yanwei and Liu, Ziwei and Li, Chunyuan},
 journal = {ArXiv preprint},
 title = {LLaVA-OneVision: Easy Visual Task Transfer},
 volume = {abs/2408.03326},
 year = {2024}
}

@inproceedings{li2024llamavid,
 author = {Li, Yanwei and Wang, Chengyao and Jia, Jiaya},
 journal = {European Conference on Computer Vision},
 title = {LLaMA-VID: An Image is Worth 2 Tokens in Large Language Models},
 year = {2024}
}

@article{shu2024video,
  title={Video-XL: Extra-Long Vision Language Model for Hour-Scale Video Understanding},
  author={Shu, Yan and Zhang, Peitian and Liu, Zheng and Qin, Minghao and Zhou, Junjie and Huang, Tiejun and Zhao, Bo},
  journal={arXiv preprint arXiv:2409.14485},
  year={2024}
}

@misc{fuyu-8b,
 author = {Bavishi, Rohan and Elsen, Erich and Hawthorne, Curtis and Nye, Maxwell and Odena, Augustus and Somani, Arushi and  Ta\c{s}\i{}rlar, Sa\u{g}nak},
 title = {Introducing our Multimodal Models},
 year = {2023}
}

@article{qwen2.5vl,
  title={Qwen2.5-VL Technical Report},
  author={Bai, Shuai and Chen, Keqin and Liu, Xuejing and Wang, Jialin and Ge, Wenbin and Song, Sibo and Dang, Kai and Wang, Peng and Wang, Shijie and Tang, Jun and Zhong, Humen and Zhu, Yuanzhi and Yang, Mingkun and Li, Zhaohai and Wan, Jianqiang and Wang, Pengfei and Ding, Wei and Fu, Zheren and Xu, Yiheng and Ye, Jiabo and Zhang, Xi and Xie, Tianbao and Cheng, Zesen and Zhang, Hang and Yang, Zhibo and Xu, Haiyang and Lin, Junyang},
  journal={arXiv preprint arXiv:2502.13923},
  year={2025}
}

@article{chen2024sharegpt4video,
 author = {Chen, Lin and Wei, Xilin and Li, Jinsong and Dong, Xiaoyi and Zhang, Pan and Zang, Yuhang and Chen, Zehui and Duan, Haodong and Lin, Bin and Tang, Zhenyu and Yuan, Li and Qiao, Yu and Lin, Dahua and Zhao, Feng and Wang, Jiaqi},
 journal = {ArXiv preprint},
 title = {ShareGPT4Video: Improving Video Understanding and Generation with Better Captions},
 volume = {abs/2406.04325},
 year = {2024}
}

@article{lin2023video,
 author = {Lin, Bin and Zhu, Bin and Ye, Yang and Ning, Munan and Jin, Peng and Yuan, Li},
 journal = {ArXiv preprint},
 title = {Video-LLaVA: Learning United Visual Representation by Alignment Before Projection},
 volume = {abs/2311.10122},
 year = {2023}
}

@misc{lin2023vila,
 archiveprefix = {arXiv},
 author = {Ji Lin and Hongxu Yin and Wei Ping and Yao Lu and Pavlo Molchanov and Andrew Tao and Huizi Mao and Jan Kautz and Mohammad Shoeybi and Song Han},
 eprint = {2312.07533},
 primaryclass = {cs.CV},
 title = {VILA: On Pre-training for Visual Language Models},
 year = {2023}
}

@inproceedings{Weng2024LongVLMEL,
  title={LongVLM: Efficient Long Video Understanding via Large Language Models},
  author={Yuetian Weng and Mingfei Han and Haoyu He and Xiaojun Chang and Bohan Zhuang},
  booktitle={European Conference on Computer Vision},
  year={2024},
  url={https://api.semanticscholar.org/CorpusID:268889590}
}

@article{zhang2024longva,
 author = {Peiyuan Zhang and Kaichen Zhang and Bo Li and Guangtao Zeng and Jingkang Yang and Yuanhan Zhang and Ziyue Wang and Haoran Tan and Chunyuan Li and Ziwei Liu},
 journal = {ArXiv preprint},
 title = {Long Context Transfer from Language to Vision},
 volume = {abs/2406.16852},
 year = {2024}
}

@article{zhang2025videollama3,
  title={VideoLLaMA 3: Frontier Multimodal Foundation Models for Image and Video Understanding},
  author={Zhang, Boqiang and Li, Kehan and Cheng, Zesen and Hu, Zhiqiang and Yuan, Yuqian and Chen, Guanzheng and Leng, Sicong and Jiang, Yuming and Zhang, Hang and Li, Xin and others},
  journal={arXiv preprint arXiv:2501.13106},
  year={2025}
}

@article{liu2024nvila,
  title={NVILA: Efficient frontier visual language models},
  author={Liu, Zhijian and Zhu, Ligeng and Shi, Baifeng and Zhang, Zhuoyang and Lou, Yuming and Yang, Shang and Xi, Haocheng and Cao, Shiyi and Gu, Yuxian and Li, Dacheng and others},
  journal={arXiv preprint arXiv:2412.04468},
  year={2024}
}

@article{wang2025internvideo2,
  title={InternVideo2. 5: Empowering Video MLLMs with Long and Rich Context Modeling},
  author={Wang, Yi and Li, Xinhao and Yan, Ziang and He, Yinan and Yu, Jiashuo and Zeng, Xiangyu and Wang, Chenting and Ma, Changlian and Huang, Haian and Gao, Jianfei and others},
  journal={arXiv preprint arXiv:2501.12386},
  year={2025}
}

@article{Zhang2024LongCT,
  title={Long Context Transfer from Language to Vision},
  author={Peiyuan Zhang and Kaichen Zhang and Bo Li and Guangtao Zeng and Jingkang Yang and Yuanhan Zhang and Ziyue Wang and Haoran Tan and Chunyuan Li and Ziwei Liu},
  journal={ArXiv},
  year={2024},
  volume={abs/2406.16852},
  url={https://api.semanticscholar.org/CorpusID:270703489}
}

@article{wang2024longllava,
  title={LongLLaVA: Scaling Multi-modal LLMs to 1000 Images Efficiently via a Hybrid Architecture},
  author={Wang, Xidong and Song, Dingjie and Chen, Shunian and Zhang, Chen and Wang, Benyou},
  journal={arXiv preprint arXiv:2409.02889},
  year={2024}
}

@article{videoccam,
  title={Video-ccam: Enhancing video-language understanding with causal cross-attention masks for short and long videos},
  author={Fei, Jiajun and Li, Dian and Deng, Zhidong and Wang, Zekun and Liu, Gang and Wang, Hui},
  journal={arXiv preprint arXiv:2408.14023},
  year={2024}
}

@article{Rafailov2023DirectPO,
  title={Direct Preference Optimization: Your Language Model is Secretly a Reward Model},
  author={Rafael Rafailov and Archit Sharma and Eric Mitchell and Stefano Ermon and Christopher D. Manning and Chelsea Finn},
  journal={ArXiv},
  year={2023},
  volume={abs/2305.18290},
  url={https://api.semanticscholar.org/CorpusID:258959321}
}

@article{videochat-flash,
  title={Videochat-flash: Hierarchical compression for long-context video modeling},
  author={Li, Xinhao and Wang, Yi and Yu, Jiashuo and Zeng, Xiangyu and Zhu, Yuhan and Huang, Haian and Gao, Jianfei and Li, Kunchang and He, Yinan and Wang, Chenting and others},
  journal={arXiv preprint arXiv:2501.00574},
  year={2024}
}

@inproceedings{Maaz2023VideoChatGPT,
 author = {Maaz, Muhammad and Rasheed, Hanoona and Khan, Salman and Khan, Fahad Shahbaz},
 booktitle = {Proceedings of the 62nd Annual Meeting of the Association for Computational Linguistics (ACL 2024)},
 title = {Video-ChatGPT: Towards Detailed Video Understanding via Large Vision and Language Models},
 year = {2024}
}

@article{MLVU,
 author = {Zhou, Junjie and Shu, Yan and Zhao, Bo and Wu, Boya and Xiao, Shitao and Yang, Xi and Xiong, Yongping and Zhang, Bo and Huang, Tiejun and Liu, Zheng},
 journal = {ArXiv preprint},
 title = {MLVU: A Comprehensive Benchmark for Multi-Task Long Video Understanding},
 volume = {abs/2406.04264},
 year = {2024}
}

@article{2020t5,
  author  = {Colin Raffel and Noam Shazeer and Adam Roberts and Katherine Lee and Sharan Narang and Michael Matena and Yanqi Zhou and Wei Li and Peter J. Liu},
  title   = {Exploring the Limits of Transfer Learning with a Unified Text-to-Text Transformer},
  journal = {Journal of Machine Learning Research},
  year    = {2020},
  volume  = {21},
  number  = {140},
  pages   = {1-67},
  url     = {http://jmlr.org/papers/v21/20-074.html}
}

@article{videomme,
 author = {Fu, Chaoyou and Dai, Yuhan and Luo, Yondong and Li, Lei and Ren, Shuhuai and Zhang, Renrui and Wang, Zihan and Zhou, Chenyu and Shen, Yunhang and Zhang, Mengdan and others},
 journal = {ArXiv preprint},
 title = {Video-MME: The First-Ever Comprehensive Evaluation Benchmark of Multi-modal LLMs in Video Analysis},
 volume = {abs/2405.21075},
 year = {2024}
}

@inproceedings{tempcompass,
 author = {Liu, Yuanxin  and
Li, Shicheng  and
Liu, Yi  and
Wang, Yuxiang  and
Ren, Shuhuai  and
Li, Lei  and
Chen, Sishuo  and
Sun, Xu  and
Hou, Lu},
 booktitle = {Findings of the Association for Computational Linguistics ACL 2024},
 editor = {Ku, Lun-Wei  and
Martins, Andre  and
Srikumar, Vivek},
 pages = {8731--8772},
 title = {{T}emp{C}ompass: Do Video {LLM}s Really Understand Videos?},
 year = {2024}
}

@article{longvila,
  title={Longvila: Scaling long-context visual language models for long videos},
  author={Xue, Fuzhao and Chen, Yukang and Li, Dacheng and Hu, Qinghao and Zhu, Ligeng and Li, Xiuyu and Fang, Yunhao and Tang, Haotian and Yang, Shang and Liu, Zhijian and others},
  journal={arXiv preprint arXiv:2408.10188},
  year={2024}
}

@inproceedings{kwon2023efficient,
  title={Efficient Memory Management for Large Language Model Serving with PagedAttention},
  author={Woosuk Kwon and Zhuohan Li and Siyuan Zhuang and Ying Sheng and Lianmin Zheng and Cody Hao Yu and Joseph E. Gonzalez and Hao Zhang and Ion Stoica},
  booktitle={Proceedings of the ACM SIGOPS 29th Symposium on Operating Systems Principles},
  year={2023}
}

@article{liu2024oryx,
  title={Oryx mllm: On-demand spatial-temporal understanding at arbitrary resolution},
  author={Liu, Zuyan and Dong, Yuhao and Liu, Ziwei and Hu, Winston and Lu, Jiwen and Rao, Yongming},
  journal={arXiv preprint arXiv:2409.12961},
  year={2024}
}

@article{wang2024tarsier,
  title={Tarsier: Recipes for training and evaluating large video description models},
  author={Wang, Jiawei and Yuan, Liping and Zhang, Yuchen and Sun, Haomiao},
  journal={arXiv preprint arXiv:2407.00634},
  year={2024}
}

@article{hurst2024gpt,
  title={Gpt-4o system card},
  author={Hurst, Aaron and Lerer, Adam and Goucher, Adam P and Perelman, Adam and Ramesh, Aditya and Clark, Aidan and Ostrow, AJ and Welihinda, Akila and Hayes, Alan and Radford, Alec and others},
  journal={arXiv preprint arXiv:2410.21276},
  year={2024}
}

@article{chen2024expanding,
  title={Expanding performance boundaries of open-source multimodal models with model, data, and test-time scaling},
  author={Chen, Zhe and Wang, Weiyun and Cao, Yue and Liu, Yangzhou and Gao, Zhangwei and Cui, Erfei and Zhu, Jinguo and Ye, Shenglong and Tian, Hao and Liu, Zhaoyang and others},
  journal={arXiv preprint arXiv:2412.05271},
  year={2024}
}

@article{hong2024cogvlm2,
  title={Cogvlm2: Visual language models for image and video understanding},
  author={Hong, Wenyi and Wang, Weihan and Ding, Ming and Yu, Wenmeng and Lv, Qingsong and Wang, Yan and Cheng, Yean and Huang, Shiyu and Ji, Junhui and Xue, Zhao and others},
  journal={arXiv preprint arXiv:2408.16500},
  year={2024}
}

@article{wang2024cogvlm,
  title={Cogvlm: Visual expert for pretrained language models},
  author={Wang, Weihan and Lv, Qingsong and Yu, Wenmeng and Hong, Wenyi and Qi, Ji and Wang, Yan and Ji, Junhui and Yang, Zhuoyi and Zhao, Lei and XiXuan, Song and others},
  journal={Advances in Neural Information Processing Systems},
  volume={37},
  pages={121475--121499},
  year={2024}
}

@article{zhang2024video,
  title={Video instruction tuning with synthetic data},
  author={Zhang, Yuanhan and Wu, Jinming and Li, Wei and Li, Bo and Ma, Zejun and Liu, Ziwei and Li, Chunyuan},
  journal={arXiv preprint arXiv:2410.02713},
  year={2024}
}

@inproceedings{hemmworld,
  title={MMWorld: Towards Multi-discipline Multi-faceted World Model Evaluation in Videos},
  author={He, Xuehai and Feng, Weixi and Zheng, Kaizhi and Lu, Yujie and Zhu, Wanrong and Li, Jiachen and Fan, Yue and Wang, Jianfeng and Li, Linjie and Yang, Zhengyuan and others},
  booktitle={The Thirteenth International Conference on Learning Representations}
}

@article{patraucean2023perception,
  title={Perception test: A diagnostic benchmark for multimodal video models},
  author={Patraucean, Viorica and Smaira, Lucas and Gupta, Ankush and Recasens, Adria and Markeeva, Larisa and Banarse, Dylan and Koppula, Skanda and Malinowski, Mateusz and Yang, Yi and Doersch, Carl and others},
  journal={Advances in Neural Information Processing Systems},
  volume={36},
  pages={42748--42761},
  year={2023}
}

@inproceedings{li2024mvbench,
  title={Mvbench: A comprehensive multi-modal video understanding benchmark},
  author={Li, Kunchang and Wang, Yali and He, Yinan and Li, Yizhuo and Wang, Yi and Liu, Yi and Wang, Zun and Xu, Jilan and Chen, Guo and Luo, Ping and others},
  booktitle={Proceedings of the IEEE/CVF Conference on Computer Vision and Pattern Recognition},
  pages={22195--22206},
  year={2024}
}

@article{zhang2024eventhallusion,
  title={Eventhallusion: Diagnosing event hallucinations in video llms},
  author={Zhang, Jiacheng and Jiao, Yang and Chen, Shaoxiang and Zhao, Na and Chen, Jingjing},
  journal={arXiv preprint arXiv:2409.16597},
  year={2024}
}

@article{liu2024playground,
  title={Playground v3: Improving text-to-image alignment with deep-fusion large language models},
  author={Liu, Bingchen and Akhgari, Ehsan and Visheratin, Alexander and Kamko, Aleks and Xu, Linmiao and Shrirao, Shivam and Lambert, Chase and Souza, Joao and Doshi, Suhail and Li, Daiqing},
  journal={arXiv preprint arXiv:2409.10695},
  year={2024}
}

@inproceedings{kembhavi2016diagram,
title={A diagram is worth a dozen images},
author={Kembhavi, Aniruddha and Salvato, Mike and Kolve, Eric and Seo, Minjoon and Hajishirzi, Hannaneh and Farhadi, Ali},
booktitle={European conference on computer vision},
pages={235--251},
year={2016},
organization={Springer}
}

@article{zhang2024lmms,
  title={Lmms-eval: Reality check on the evaluation of large multimodal models},
  author={Zhang, Kaichen and Li, Bo and Zhang, Peiyuan and Pu, Fanyi and Cahyono, Joshua Adrian and Hu, Kairui and Liu, Shuai and Zhang, Yuanhan and Yang, Jingkang and Li, Chunyuan and others},
  journal={arXiv preprint arXiv:2407.12772},
  year={2024}
}

@article{shi2025slow,
  title={Slow-Fast Architecture for Video Multi-Modal Large Language Models},
  author={Shi, Min and Wang, Shihao and Chen, Chieh-Yun and Jain, Jitesh and Wang, Kai and Xiong, Junjun and Liu, Guilin and Yu, Zhiding and Shi, Humphrey},
  journal={arXiv preprint arXiv:2504.01328},
  year={2025}
}

@inproceedings{sharma2018conceptual,
  title={Conceptual captions: A cleaned, hypernymed, image alt-text dataset for automatic image captioning},
  author={Sharma, Piyush and Ding, Nan and Goodman, Sebastian and Soricut, Radu},
  booktitle={Proceedings of the 56th Annual Meeting of the Association for Computational Linguistics (Volume 1: Long Papers)},
  pages={2556--2565},
  year={2018}
}

@article{ordonez2011im2text,
  title={Im2text: Describing images using 1 million captioned photographs},
  author={Ordonez, Vicente and Kulkarni, Girish and Berg, Tamara},
  journal={Advances in neural information processing systems},
  volume={24},
  year={2011}
}

@misc{coyo-700m,
  title         = {COYO-700M: Image-text pair dataset},
  author        = {Byeon, Minwoo and Park, Beomhee and Kim, Haecheon and Lee, Sungjun and Baek, Woonhyuk and Kim, Saehoon},
  year          = {2022},
  howpublished  = {\url{https://github.com/kakaobrain/coyo-dataset}},
}

@misc{lacion-coco,
  title         = {Laion coco: 600m synthetic captions from laion2b-en.},
  author        = {},
  howpublished  = {\url{https://laion.ai/blog/laion-coco/}},
}

@article{gu2022wukong,
  title={Wukong: A 100 million large-scale chinese cross-modal pre-training benchmark},
  author={Gu, Jiaxi and Meng, Xiaojun and Lu, Guansong and Hou, Lu and Minzhe, Niu and Liang, Xiaodan and Yao, Lewei and Huang, Runhui and Zhang, Wei and Jiang, Xin and others},
  journal={Advances in Neural Information Processing Systems},
  volume={35},
  pages={26418--26431},
  year={2022}
}

@article{wang2023internvid,
  title={Internvid: A large-scale video-text dataset for multimodal understanding and generation},
  author={Wang, Yi and He, Yinan and Li, Yizhuo and Li, Kunchang and Yu, Jiashuo and Ma, Xin and Li, Xinhao and Chen, Guo and Chen, Xinyuan and Wang, Yaohui and others},
  journal={arXiv preprint arXiv:2307.06942},
  year={2023}
}

@inproceedings{chen2024panda,
  title={Panda-70m: Captioning 70m videos with multiple cross-modality teachers},
  author={Chen, Tsai-Shien and Siarohin, Aliaksandr and Menapace, Willi and Deyneka, Ekaterina and Chao, Hsiang-wei and Jeon, Byung Eun and Fang, Yuwei and Lee, Hsin-Ying and Ren, Jian and Yang, Ming-Hsuan and others},
  booktitle={Proceedings of the IEEE/CVF Conference on Computer Vision and Pattern Recognition},
  pages={13320--13331},
  year={2024}
}

@inproceedings{xu2022metmeme,
  title={Met-meme: A multimodal meme dataset rich in metaphors},
  author={Xu, Bo and Li, Tingting and Zheng, Junzhe and Naseriparsa, Mehdi and Zhao, Zhehuan and Lin, Hongfei and Xia, Feng},
  booktitle={Proceedings of the 45th international ACM SIGIR conference on research and development in information retrieval},
  pages={2887--2899},
  year={2022}
}

@article{meyer2024public,
  title={Public Domain 12M: A Highly Aesthetic Image-Text Dataset with Novel Governance Mechanisms},
  author={Meyer, Jordan and Padgett, Nick and Miller, Cullen and Exline, Laura},
  journal={arXiv preprint arXiv:2410.23144},
  year={2024}
}

@article{hsieh2024graph,
  title={Graph-Based Captioning: Enhancing Visual Descriptions by Interconnecting Region Captions},
  author={Hsieh, Yu-Guan and Hsieh, Cheng-Yu and Yeh, Shih-Ying and B{\'e}thune, Louis and Ansari, Hadi Pour and Vasu, Pavan Kumar Anasosalu and Li, Chun-Liang and Krishna, Ranjay and Tuzel, Oncel and Cuturi, Marco},
  journal={arXiv preprint arXiv:2407.06723},
  year={2024}
}

@article{li2024densefusion,
  title={Densefusion-1m: Merging vision experts for comprehensive multimodal perception},
  author={Li, Xiaotong and Zhang, Fan and Diao, Haiwen and Wang, Yueze and Wang, Xinlong and DUAN, LINGYU},
  journal={Advances in Neural Information Processing Systems},
  volume={37},
  pages={18535--18556},
  year={2024}
}

@InProceedings{Dwibedi_2020_CVPR,

author = {Dwibedi, Debidatta and Aytar, Yusuf and Tompson, Jonathan and Sermanet, Pierre and Zisserman, Andrew},

title = {Counting Out Time: Class Agnostic Video Repetition Counting in the Wild},

booktitle = {IEEE/CVF Conference on Computer Vision and Pattern Recognition (CVPR)},

month = {June},

year = {2020}

}

@article{maaz2024videogpt+,
  title={Videogpt+: Integrating image and video encoders for enhanced video understanding},
  author={Maaz, Muhammad and Rasheed, Hanoona and Khan, Salman and Khan, Fahad},
  journal={arXiv preprint arXiv:2406.09418},
  year={2024}
}

@inproceedings{toh2023mermaid,
  title={MERMAID: A Dataset and Framework for Multimodal Meme Semantic Understanding},
  author={Toh, Shaun and Kuek, Adriel and Chong, Wen-Haw and Lee, Roy Ka-Wei},
  booktitle={2023 IEEE International Conference on Big Data (BigData)},
  pages={433--442},
  year={2023},
  organization={IEEE}
}

@article{toshniwal2024openmathinstruct,
  title={Openmathinstruct-2: Accelerating ai for math with massive open-source instruction data},
  author={Toshniwal, Shubham and Du, Wei and Moshkov, Ivan and Kisacanin, Branislav and Ayrapetyan, Alexan and Gitman, Igor},
  journal={arXiv preprint arXiv:2410.01560},
  year={2024}
}

@article{monfort2021multi,
  title={Multi-moments in time: Learning and interpreting models for multi-action video understanding},
  author={Monfort, Mathew and Pan, Bowen and Ramakrishnan, Kandan and Andonian, Alex and McNamara, Barry A and Lascelles, Alex and Fan, Quanfu and Gutfreund, Dan and Feris, Rog{\'e}rio Schmidt and Oliva, Aude},
  journal={IEEE Transactions on Pattern Analysis and Machine Intelligence},
  volume={44},
  number={12},
  pages={9434--9445},
  year={2021},
  publisher={IEEE}
}

@article{hu2022transrac,
title={TransRAC: Encoding Multi-scale Temporal Correlation with Transformers for Repetitive Action Counting},
author={Hu, Huazhang and Dong, Sixun and Zhao, Yiqun and Lian, Dongze and Li, Zhengxin and Gao, Shenghua},
journal={arXiv preprint arXiv:2204.01018},
year={2022}
}

@inproceedings{yu2023celebv,
  title={Celebv-text: A large-scale facial text-video dataset},
  author={Yu, Jianhui and Zhu, Hao and Jiang, Liming and Loy, Chen Change and Cai, Weidong and Wu, Wayne},
  booktitle={Proceedings of the IEEE/CVF Conference on Computer Vision and Pattern Recognition},
  pages={14805--14814},
  year={2023}
}

@article{yuan2024chronomagic,
  title={Chronomagic-bench: A benchmark for metamorphic evaluation of text-to-time-lapse video generation},
  author={Yuan, Shenghai and Huang, Jinfa and Xu, Yongqi and Liu, Yaoyang and Zhang, Shaofeng and Shi, Yujun and Zhu, Rui-Jie and Cheng, Xinhua and Luo, Jiebo and Yuan, Li},
  journal={Advances in Neural Information Processing Systems},
  volume={37},
  pages={21236--21270},
  year={2024}
}

@article{tan2024vidgen,
  title={Vidgen-1m: A large-scale dataset for text-to-video generation},
  author={Tan, Zhiyu and Yang, Xiaomeng and Qin, Luozheng and Li, Hao},
  journal={arXiv preprint arXiv:2408.02629},
  year={2024}
}

@article{chen2024compcap,
  title={CompCap: Improving Multimodal Large Language Models with Composite Captions},
  author={Chen, Xiaohui and Shukla, Satya Narayan and Azab, Mahmoud and Singh, Aashu and Wang, Qifan and Yang, David and Peng, ShengYun and Yu, Hanchao and Yan, Shen and Zhang, Xuewen and others},
  journal={arXiv preprint arXiv:2412.05243},
  year={2024}
}

@inproceedings{zhou2018towards,
  title={Towards automatic learning of procedures from web instructional videos},
  author={Zhou, Luowei and Xu, Chenliang and Corso, Jason},
  booktitle={Proceedings of the AAAI conference on artificial intelligence},
  volume={32},
  number={1},
  year={2018}
}

@article{yu2023metamath,
  title={Metamath: Bootstrap your own mathematical questions for large language models},
  author={Yu, Longhui and Jiang, Weisen and Shi, Han and Yu, Jincheng and Liu, Zhengying and Zhang, Yu and Kwok, James T and Li, Zhenguo and Weller, Adrian and Liu, Weiyang},
  journal={arXiv preprint arXiv:2309.12284},
  year={2023}
}

@article{tuo2023anytext,
  title={Anytext: Multilingual visual text generation and editing},
  author={Tuo, Yuxiang and Xiang, Wangmeng and He, Jun-Yan and Geng, Yifeng and Xie, Xuansong},
  journal={arXiv preprint arXiv:2311.03054},
  year={2023}
}

@article{taesiri2024videogamebunny,
  title={VideoGameBunny: Towards vision assistants for video games},
  author={Taesiri, Mohammad Reza and Bezemer, Cor-Paul},
  journal={arXiv preprint arXiv:2407.15295},
  year={2024}
}

@article{ju2024miradata,
  title={Miradata: A large-scale video dataset with long durations and structured captions},
  author={Ju, Xuan and Gao, Yiming and Zhang, Zhaoyang and Yuan, Ziyang and Wang, Xintao and Zeng, Ailing and Xiong, Yu and Xu, Qiang and Shan, Ying},
  journal={Advances in Neural Information Processing Systems},
  volume={37},
  pages={48955--48970},
  year={2024}
}

@inproceedings{zhao2019hacs,
  title={Hacs: Human action clips and segments dataset for recognition and temporal localization},
  author={Zhao, Hang and Torralba, Antonio and Torresani, Lorenzo and Yan, Zhicheng},
  booktitle={Proceedings of the IEEE International Conference on Computer Vision},
  pages={8668--8678},
  year={2019}
}

@article{nan2024openvid,
  title={OpenVid-1M: A Large-Scale High-Quality Dataset for Text-to-video Generation},
  author={Nan, Kepan and Xie, Rui and Zhou, Penghao and Fan, Tiehan and Yang, Zhenheng and Chen, Zhijie and Li, Xiang and Yang, Jian and Tai, Ying},
  journal={arXiv preprint arXiv:2407.02371},
  year={2024}
}

@article{carreira2019short,
  title={A short note on the kinetics-700 human action dataset},
  author={Carreira, Joao and Noland, Eric and Hillier, Chloe and Zisserman, Andrew},
  journal={arXiv preprint arXiv:1907.06987},
  year={2019}
}

@article{liu2024mmdu,
  title={MMDU: A Multi-Turn Multi-Image Dialog Understanding Benchmark and Instruction-Tuning Dataset for LVLMs},
  author={Liu, Ziyu and Chu, Tao and Zang, Yuhang and Wei, Xilin and Dong, Xiaoyi and Zhang, Pan and Liang, Zijian and Xiong, Yuanjun and Qiao, Yu and Lin, Dahua and others},
  journal={arXiv preprint arXiv:2406.11833},
  year={2024}
}

@ONLINE{wikidump,
    author = "Wikimedia Foundation",
    title  = "Wikimedia Downloads",
    url    = "https://dumps.wikimedia.org"
}

@misc{laurençon2024unlocking,
      title={Unlocking the conversion of Web Screenshots into HTML Code with the WebSight Dataset}, 
      author={Hugo Laurençon and Léo Tronchon and Victor Sanh},
      year={2024},
      eprint={2403.09029},
      archivePrefix={arXiv},
      primaryClass={cs.HC}
}

@article{Kosmos2,
  title={Kosmos-2: Grounding Multimodal Large Language Models to the World},
  author={Zhiliang Peng and Wenhui Wang and Li Dong and Yaru Hao and Shaohan Huang and Shuming Ma and Furu Wei},
  journal={ArXiv},
  year={2023},
  volume={abs/2306.14824}
}

@article{liu2024mminstruct,
  title={Mminstruct: A high-quality multi-modal instruction tuning dataset with extensive diversity},
  author={Liu, Yangzhou and Cao, Yue and Gao, Zhangwei and Wang, Weiyun and Chen, Zhe and Wang, Wenhai and Tian, Hao and Lu, Lewei and Zhu, Xizhou and Lu, Tong and others},
  journal={Science China Information Sciences},
  volume={67},
  number={12},
  pages={1--16},
  year={2024},
  publisher={Springer}
}

@article{xiong2024lvd,
  title={Lvd-2m: A long-take video dataset with temporally dense captions},
  author={Xiong, Tianwei and Wang, Yuqing and Zhou, Daquan and Lin, Zhijie and Feng, Jiashi and Liu, Xihui},
  journal={arXiv preprint arXiv:2410.10816},
  year={2024}
}

@article{luo2024mmevol,
  title={Mmevol: Empowering multimodal large language models with evol-instruct},
  author={Luo, Run and Zhang, Haonan and Chen, Longze and Lin, Ting-En and Liu, Xiong and Wu, Yuchuan and Yang, Min and Wang, Minzheng and Zeng, Pengpeng and Gao, Lianli and others},
  journal={arXiv preprint arXiv:2409.05840},
  year={2024}
}

@ONLINE{dalle3,
    author = "Nagengast, Zach and Pach, Eduardo and Maltsev, Seva and Egan, Ben",
    title  = "Dataset Card for LAION DALL·E 3 Discord Dataset",
    url    = "https://huggingface.co/datasets/OpenDatasets/dalle-3-dataset"
}

@ONLINE{omegaproject,
    author = "OMEGA Lab",
    title  = "OMEGA Labs Bittensor Subnet: Multimodal Dataset for AGI Research",
    url    = "https://github.com/omegalabsinc/omegalabs-bittensor-subnet/tree/main"
}

@inproceedings{bain2021frozen,
  title={Frozen in time: A joint video and image encoder for end-to-end retrieval},
  author={Bain, Max and Nagrani, Arsha and Varol, G{\"u}l and Zisserman, Andrew},
  booktitle={Proceedings of the IEEE/CVF international conference on computer vision},
  pages={1728--1738},
  year={2021}
}

@article{epstein2019oops,
title={Oops! Predicting Unintentional Action in Video},
author={Epstein, Dave and Chen, Boyuan and Vondrick, Carl.},
journal={arXiv preprint arXiv:1911.11206},
year={2019}
}

@inproceedings{lu2024kvq,
  title={Kvq: Kwai video quality assessment for short-form videos},
  author={Lu, Yiting and Li, Xin and Pei, Yajing and Yuan, Kun and Xie, Qizhi and Qu, Yunpeng and Sun, Ming and Zhou, Chao and Chen, Zhibo},
  booktitle={Proceedings of the IEEE/CVF Conference on Computer Vision and Pattern Recognition},
  pages={25963--25973},
  year={2024}
}

@article{wang2024pin,
  title={Pin: A knowledge-intensive dataset for paired and interleaved multimodal documents},
  author={Wang, Junjie and Zhang, Yin and Ji, Yatai and Zhang, Yuxiang and Jiang, Chunyang and Wang, Yubo and Zhu, Kang and Wang, Zekun and Wang, Tiezhen and Huang, Wenhao and others},
  journal={arXiv preprint arXiv:2406.13923},
  year={2024}
}

@article{rohrbach2017movie,
  title={Movie description},
  author={Rohrbach, Anna and Torabi, Atousa and Rohrbach, Marcus and Tandon, Niket and Pal, Christopher and Larochelle, Hugo and Courville, Aaron and Schiele, Bernt},
  journal={International Journal of Computer Vision},
  volume={123},
  pages={94--120},
  year={2017},
  publisher={Springer}
}

@inproceedings{monfort2021spoken,
  title={Spoken moments: Learning joint audio-visual representations from video descriptions},
  author={Monfort, Mathew and Jin, SouYoung and Liu, Alexander and Harwath, David and Feris, Rogerio and Glass, James and Oliva, Aude},
  booktitle={Proceedings of the IEEE/CVF Conference on Computer Vision and Pattern Recognition},
  pages={14871--14881},
  year={2021}
}

@inproceedings{laurenccon2024building,
  title={Building and better understanding vision-language models: insights and future directions},
  author={Lauren{\c{c}}on, Hugo and Marafioti, Andr{\'e}s and Sanh, Victor and Tronchon, L{\'e}o},
  booktitle={Workshop on Responsibly Building the Next Generation of Multimodal Foundational Models},
  year={2024}
}

@inproceedings{sun2019icdar,
  title={ICDAR 2019 competition on large-scale street view text with partial labeling-RRC-LSVT},
  author={Sun, Yipeng and Ni, Zihan and Chng, Chee-Kheng and Liu, Yuliang and Luo, Canjie and Ng, Chun Chet and Han, Junyu and Ding, Errui and Liu, Jingtuo and Karatzas, Dimosthenis and others},
  booktitle={2019 International Conference on Document Analysis and Recognition (ICDAR)},
  pages={1557--1562},
  year={2019},
  organization={IEEE}
}

@inproceedings{chng2019icdar2019,
  title={Icdar2019 robust reading challenge on arbitrary-shaped text-rrc-art},
  author={Chng, Chee Kheng and Liu, Yuliang and Sun, Yipeng and Ng, Chun Chet and Luo, Canjie and Ni, Zihan and Fang, ChuanMing and Zhang, Shuaitao and Han, Junyu and Ding, Errui and others},
  booktitle={2019 International Conference on Document Analysis and Recognition (ICDAR)},
  pages={1571--1576},
  year={2019},
  organization={IEEE}
}

@article{luo2024video,
  title={Video-RAG: Visually-aligned Retrieval-Augmented Long Video Comprehension},
  author={Luo, Yongdong and Zheng, Xiawu and Yang, Xiao and Li, Guilin and Lin, Haojia and Huang, Jinfa and Ji, Jiayi and Chao, Fei and Luo, Jiebo and Ji, Rongrong},
  journal={arXiv preprint arXiv:2411.13093},
  year={2024}
}

@misc{deepseekai2025deepseekr1,
      title={DeepSeek-R1: Incentivizing Reasoning Capability in LLMs via Reinforcement Learning}, 
      author={DeepSeek-AI},
      year={2025},
      eprint={2501.12948},
      archivePrefix={arXiv},
      primaryClass={cs.CL},
      url={https://arxiv.org/abs/2501.12948}, 
}

@article{wang2024grounded,
  title={Grounded-videollm: Sharpening fine-grained temporal grounding in video large language models},
  author={Wang, Haibo and Xu, Zhiyang and Cheng, Yu and Diao, Shizhe and Zhou, Yufan and Cao, Yixin and Wang, Qifan and Ge, Weifeng and Huang, Lifu},
  journal={arXiv preprint arXiv:2410.03290},
  year={2024}
}

@article{deepseek_vl2,
  title={Deepseek-vl2: Mixture-of-experts vision-language models for advanced multimodal understanding},
  author={Wu, Zhiyu and Chen, Xiaokang and Pan, Zizheng and Liu, Xingchao and Liu, Wen and Dai, Damai and Gao, Huazuo and Ma, Yiyang and Wu, Chengyue and Wang, Bingxuan and others},
  journal={arXiv preprint arXiv:2412.10302},
  year={2024}
}

@article{yang2024vript,
  title={Vript: A video is worth thousands of words},
  author={Yang, Dongjie and Huang, Suyuan and Lu, Chengqiang and Han, Xiaodong and Zhang, Haoxin and Gao, Yan and Hu, Yao and Zhao, Hai},
  journal={Advances in Neural Information Processing Systems},
  volume={37},
  pages={57240--57261},
  year={2024}
}

@article{srinivasan2021wit,
  title={WIT: Wikipedia-based Image Text Dataset for Multimodal Multilingual Machine Learning},
  author={Srinivasan, Krishna and Raman, Karthik and Chen, Jiecao and Bendersky, Michael and Najork, Marc},
  journal={arXiv preprint arXiv:2103.01913},
  year={2021}
}

@article{awadalla2024blip3,
  title={BLIP3-KALE: Knowledge Augmented Large-Scale Dense Captions},
  author={Awadalla, Anas and Xue, Le and Shu, Manli and Yan, An and Wang, Jun and Purushwalkam, Senthil and Shen, Sheng and Lee, Hannah and Lo, Oscar and Park, Jae Sung and others},
  journal={arXiv preprint arXiv:2411.07461},
  year={2024}
}

@inproceedings{wang2024all,
  title={The all-seeing project v2: Towards general relation comprehension of the open world},
  author={Wang, Weiyun and Ren, Yiming and Luo, Haowen and Li, Tiantong and Yan, Chenxiang and Chen, Zhe and Wang, Wenhai and Li, Qingyun and Lu, Lewei and Zhu, Xizhou and others},
  booktitle={European Conference on Computer Vision},
  pages={471--490},
  year={2024},
  organization={Springer}
}

@article{gu2024infinity,
  title={Infinity-mm: Scaling multimodal performance with large-scale and high-quality instruction data},
  author={Gu, Shuhao and Zhang, Jialing and Zhou, Siyuan and Yu, Kevin and Xing, Zhaohu and Wang, Liangdong and Cao, Zhou and Jia, Jintao and Zhang, Zhuoyi and Wang, Yixuan and others},
  journal={arXiv preprint arXiv:2410.18558},
  year={2024}
}

@article{Jiang2024MANTISIM,
  title={MANTIS: Interleaved Multi-Image Instruction Tuning},
  author={Dongfu Jiang and Xuan He and Huaye Zeng and Cong Wei and Max W.F. Ku and Qian Liu and Wenhu Chen},
  journal={Transactions on Machine Learning Research},
  year={2024},
  volume={2024},
  url={https://openreview.net/forum?id=skLtdUVaJa}
}

@misc{han2023shot2story20k,
    title={Shot2Story20K: A New Benchmark for Comprehensive Understanding of Multi-shot Videos}, 
    author={Mingfei Han and Linjie Yang and Xiaojun Chang and Heng Wang},
    year={2023},
    eprint={2312.10300},
    archivePrefix={arXiv},
    primaryClass={cs.CV}
}

@misc{sharegemini,
  title={ShareGemini: Scaling Up Video Caption Data for Multimodal Large Language Models},
  url={https://github.com/Share14/ShareGemini},
  author={Share},
  month={June},
  year={2024}
}

@article{rawal2024cinepile,
  title={Cinepile: A long video question answering dataset and benchmark},
  author={Rawal, Ruchit and Saifullah, Khalid and Farr{\'e}, Miquel and Basri, Ronen and Jacobs, David and Somepalli, Gowthami and Goldstein, Tom},
  journal={arXiv preprint arXiv:2405.08813},
  year={2024}
}
